%% file: main.tex
\documentclass{article}

\input{preamble}

\usepackage{amsmath,bm}
\usepackage[normalem]{ulem}
\usepackage{wrapfig}
\usepackage{subcaption}
\usepackage{bookmark}

\title{Denoising Multi-$\beta$ VAE: Representation Learning for \\ Disentanglement and Generation}

\author{%
  Anshuk Uppal\thanks{Work done during an internship at Sony AI.}\\
  Technical University of Denmark,\\
  Copenhagen, Denmark\\
  \texttt{ansup@dtu.dk} \\
\And
Yuhta Takida\\
Sony AI\\
Tokyo, Japan\\
\And
Chieh-Hsin Lai\\
Sony AI\\
Tokyo, Japan\\
\And
Yuki Mitsufuji\\
Sony AI \& Sony Group Corporation\\
New York, USA\\
}

\definecolor{c_cm}{rgb}{0.6,0.6,0.6}
\definecolor{c_yuhta}{rgb}{0.0, 0.0, 0.0}
\newcommand{\cyuhta}{\color{black}} %
\newcommand{\yuhta}[1]{{\color{black}{{#1}}}} %

\definecolor{c_ansh}{rgb}{0.0, 0.72, 0.92}

\begin{document}

\maketitle

\input{sec/0_abstract}

\input{proposed_sec/1_intro}

\input{proposed_sec/2_overview}
\input{proposed_sec/3_conditionalvae}

\input{proposed_sec/3_ldm}
\input{proposed_sec/4_relatedworks}

\input{sec/5_experiments}

\input{sec/6_futurework}

\begin{refcontext}[sorting=nyt]
\printbibliography[heading=bibliography]
\end{refcontext}

\appendix
\newpage
\tableofcontents

\input{app/beta_vae}

\input{app/proofs}
\input{app/experiment_deets}

\input{app/extra_results}
\input{app/limitations}

\end{document}

%% file: preamble.tex
\PassOptionsToPackage{round}{natbib}
\usepackage[preprint,nonatbib]{neurips_2025}  

\usepackage[utf8]{inputenc} %
\usepackage[T1]{fontenc}    %
\usepackage[breaklinks,colorlinks]{hyperref}%

\usepackage{url}            %
\usepackage{booktabs}       %
\usepackage{amsfonts}       %
\usepackage{nicefrac}       %
\usepackage{microtype}      %
\usepackage[dvipsnames,svgnames,x11names]{xcolor}         %
\usepackage{wrapfig}
\usepackage{enumitem}

\usepackage{amsmath}
\usepackage{amssymb}
\usepackage{mathtools}
\usepackage{adjustbox}
\usepackage{amsthm}
\usepackage{threeparttable}

\usepackage{xpatch}                     %
\usepackage[backend=biber,
            style=authoryear-comp,
            uniquelist=false,           %
            maxnames=100,               %
            maxcitenames=1,             %
            sortcites=false,            %
            natbib=true,
            isbn=false,
            url=false,
            doi=false,
            eprint=false,
            ]{biblatex}

\usepackage[ruled,vlined]{algorithm2e}
\usepackage[breaklinks,colorlinks]{hyperref}%
\usepackage[capitalise]{cleveref}

\usepackage{graphicx}
\usepackage{subcaption}
\usepackage{array}
\usepackage{siunitx}
\usepackage{soul}
\usepackage{commath}
\usepackage{xspace}
\usepackage{multirow, multicol}
\usepackage{fixme}

\graphicspath{{./figuresConcrete/}}
\definecolor{dark-red}{rgb}{0.4,0.15,0.15}
\definecolor{dark-blue}{rgb}{0.15,0.15,0.4}
\definecolor{medium-blue}{rgb}{0,0,0.5}
\definecolor{light-grey}{rgb}{0.98,0.98,0.98}
\colorlet{light-blue}{blue!50!cyan}
\colorlet{light-red}{red!50!purple}

\hypersetup{
    colorlinks,
    linkcolor={Firebrick3},
    citecolor={light-blue},
    urlcolor={light-blue},
    unicode=true,
    pdfcreator  = \LaTeXe,%
    pdfproducer = \LaTeXe,%
}

\theoremstyle{plain}
\newtheorem{theorem}{Theorem}[section]
\newtheorem{proposition}[theorem]{Proposition}

\theoremstyle{definition}

\theoremstyle{remark}

\DeclareMathOperator{\KL}{D_{KL}}

\renewcommand{\vec}[1]{\boldsymbol{#1}}
\newcommand{\mat}[1]{\boldsymbol{#1}}

\newcommand\kldiv[1][]{\:#1\|\:}

\DeclarePairedDelimiterX\KLparen[1][]{\let\to\skldiv #1}

\newcommand{\E}{\mathbb{E}}                     %

\newcommand{\vf}{\vec{f}}
\newcommand{\vg}{\vec{g}}

\newcommand{\vx}{\vec{x}}

\newcommand{\vz}{\vec{z}}

\newcommand{\mI}{\mat{I}}

\NewDocumentCommand{\codeword}{v}{%
\texttt{\textcolor{blue}{#1}}%
}

%% file: sec/0_abstract.tex
\begin{abstract}
\color{black}

Disentangled and interpretable latent representations in generative models typically come at the cost of generation quality. The $\beta$-VAE framework introduces a hyperparameter $\beta$ to balance disentanglement and reconstruction quality, where setting $\beta > 1$ introduces an information bottleneck that favors disentanglement over sharp, accurate reconstructions. To address this trade-off, we propose a novel generative modeling framework that leverages a range of $\beta$ values to learn multiple corresponding latent representations. First, we obtain a slew of representations by training a single variational autoencoder (VAE), with a new loss function that controls the information retained in each latent representation such that the higher $\beta$ value prioritize disentanglement over reconstruction fidelity. We then, introduce a non-linear diffusion model that smoothly transitions latent representations corresponding to different $\beta$ values. This model denoises towards less disentangled and more informative representations, ultimately leading to (almost) lossless representations, enabling sharp reconstructions. Furthermore, our model supports sample generation without input images, functioning as a standalone generative model. We evaluate our framework in terms of both disentanglement and generation quality. Additionally, we observe smooth transitions in the latent spaces with respect to changes in $\beta$, facilitating consistent manipulation of generated outputs.

\color{black}

\end{abstract}

%% file: proposed_sec/1_intro.tex
\setcounter{section}{0}

\section{Introduction}
\label{sec:intro}

Today, numerous advanced latent generative models are capable of producing hyperrealistic images, providing end users with a broad array of options. Current advancements in state-of-the-art generative models focus primarily on qualitative improvements in generated outputs, with recent research emphasizing the study and analysis of training dynamics to enhance generation quality~\citep{Karras2022edm,Karras2024edm2,pmlr-v202-hoogeboom23a,hoogeboom2024simplerdiffusionsid215}. Consequently, progress in generative modeling has largely shifted focus from learning and evaluating a model’s latent representations to refining the generation process itself. However, research on deep latent generative models~\citep{Kingma2013AutoEncodingVB,Radford2015UnsupervisedRL} and unsupervised representation learning has shown that purposefully learned representations not only enhance generative performance but also offer practical advantages {\color{c_yuhta} such as attribute and object changes}~\citep{wu2023uncovering}.

Previous generative modeling approaches aimed at learning disentangled and interpretable latent representations have often trailed behind in generation quality. $\beta$-VAE is a fundamental method for learning such representations, based on the variational autoencoder (VAE) framework~\citep{Kingma2013AutoEncodingVB}. \citet{Higgins2016betaVAE} modified the VAE objective by introducing a hyperparameter $\beta$, where setting $\beta=1$ recovers the original objective function. This $\beta$ parameter governs the degree of disentanglement, balancing it against reconstruction and generation quality. A larger $\beta$ value imposes stronger regularization on the latent space, empirically shown to promote disentanglement, while a smaller $\beta$ prioritizes reconstruction accuracy but does not encourage disentanglement. Although $\beta$-VAE has been extensively studied~\citep{Kim2018DisentanglingBF,chen2018isolating,controlVAEshao20b,Dewangandeepconv22}, overcoming this challenging trade-off in general models remains difficult and has only been addressed by a few works~\citep{ren2022DisCo,yang2023disdiff,yang23infodiff}.

Inspired by $\beta$-VAEs, we aim to promote disentangled representation learning within modern generative models. To this end, we propose a novel generative modeling framework. Our model consists of two main components, {\color{c_yuhta}trained in a two-stage manner}. First, we train a \emph{single} VAE that learns a spectrum of latent representations by varying the parameter $\beta$, which controls disentanglement through regularization. This VAE comprises an encoder and a decoder, each conditioned on $\beta$. However, this VAE still faces the trade-off issue: disentangled latent representations with larger $\beta$ lose information about the original input, leading to blurred reconstructions similar to those of a standard $\beta$-VAE. To address this, we introduce a novel non-linear diffusion model that denoises the latent variable at a given $\beta$ back to an (ideally) non-lossy latent space corresponding to $\beta=0$. %
This allows us to generate sharp, non-blurred images by decoding the denoised latent variable (see \cref{fig:architecture}).

In our experiments, we evaluate our model in terms of both disentanglement and generation quality. For disentanglement, we demonstrate that our model effectively achieves this purpose while maintaining generation performance, by {\color{c_yuhta}following a well-established benchmark using CelebA~\citep{yang2023disdiff}}. Additionally, we benchmark our approach on well-known toy datasets
~\citep{pmlr-v97-locatello19a,Khrulkov2021DisentangledRF}. Furthermore, we test our model's capability as a standalone generative model on widely-used image datasets both qualitatively and quantitatively.
We also show that the set of learned latent spaces is smooth with respect to $\beta$, which is essential for consistent manipulation.

Our contributions are briefly listed as follows.
\begin{itemize}[leftmargin=20pt]
    \item We propose a generative modeling framework that leverages multiple levels of latent representations, ranging from fully-informed to fully-disentangled, by extending $\beta$-VAE with a range of $\beta$ values, along with a novel model design and objective function.
    \item We propose a novel non-linear diffusion model that connects latent spaces corresponding to different $\beta$ values. By combining the VAE and the diffusion model, our approach enables both disentanglement and high-quality generation in principle.
    \item We empirically demonstrate that our model effectively balances disentanglement and image quality, achieving a superior trade-off compared to existing methods with the same motivation. Our approach attains disentanglement performance on par with disentanglement-focused baselines while generating high-quality images comparable to state-of-the-art generative models.
\end{itemize}

\color{black}

%% file: proposed_sec/2_overview.tex
\setcounter{section}{1}

\section{Overview of $\beta$-VAE}
\label{sec:betaVAE}

We begin with the formulation of a vanilla VAE~\citep{Kingma2013AutoEncodingVB}. Suppose we have a dataset $\mathcal{D} = \{\vec{x}_i\}_{i=1}^{M}$, where $\vec{x}_i \in \mathbb{R}^D$ for $i = \{1,\ldots, M\}$.%
We denote the empirical distribution defined by $\mathcal{D}$ as $p_{\mathcal{D}}(\vec{x})$. A VAE aims to uncover a reduced set of latent factors that give rise to this dataset.

Specifically, a latent variable $\vec{z} \in \mathbb{R}^d$ ($d<D$) is introduced, with its prior distribution set as $p(\vec{z}) = \mathcal{N}(\mathbf{0},\mI_d)$. Data samples are generated by first sampling $\vec{z}\sim p(\vec{z})$ and then decoding it using a probabilistic decoder, denoted as $p_\theta(\vec{x}|\vec{z})$. The decoder is commonly parameterized by a conditional isotropic Gaussian as $p_\theta(\vec{x}|\vec{z}) = \mathcal{N}(\vec{x}|g_\theta(\vec{z}),s^2\mI_D)$ with a non-negative scalar $s^2$ and a function $g_\theta:\mathbb{R}^d\to\mathbb{R}^D$.
We then wish to maximize the marginal log-likelihood $\mathbb{E}_{p_{\mathcal{D}}(x)}\left[\log p_\theta(\vec{x})\right]$, where $p_\theta(\vec{x}) = \mathbb{E}_{p(\vec{z})}[p_\theta(\vec{x}|\vec{z})]$. However, this maximization is not tractable. Therefore, in the VAE framework, a surrogate objective function called the evidence lower bound (ELBO) is maximized instead, formulated as 
\begin{align}
    \log p_\theta(\vec{x})\geq\mathbb{E}_{q_\phi(\vec{z}|\vec{x})}\left[\log p_\theta(\vec{x}|\vec{z})\right] - 
    \KL(q_\phi(\vec{z}|\vec{x})||p(\vec{z})),
    \label{eq:elbo}
\end{align}
where $q_\phi(\vec{z}|\vec{x})$ is a variational distribution used to approximate the posterior distribution $p_\theta(\vec{z}|\vec{x})$, a.k.a., the encoder. A common way to model this distribution is using a conditional Gaussian as $q_\phi(\vec{z}|\vec{x}) = \mathcal{N}(\vec{z}|\vf_\phi(\vec{x}), \mathrm{diag}(\vec{\sigma}_\phi(\vec{x})))$, with functions $\vf_\phi:\mathbb{R}^D\to\mathbb{R}^d$ and $\vec{\sigma}_\phi:\mathbb{R}^D\to\mathbb{R}_{\geq0}^d$.

Through the maximization of Eq.~\eqref{eq:elbo}, the encoder learns to recover latent generative factors from the dataset, while the decoder attempts to reconstruct $\vec{x}$ from $\vec{z}$ as accurately as possible. In other words, the encoder and decoder are trained to compress the data without information loss, effectively becoming stochastic inverses of each other. The $\beta$-VAE \citep{Higgins2016betaVAE} is a variant of the above model that employs the following modified objective function: $\mathbb{E}_{q_\phi(\vec{z}|\vec{x})}\left[\log p_\theta(\vec{x}|\vec{z})\right] - \beta\KL(q_\phi(\vec{z}|\vec{x})||p(\vec{z})),$
where the \emph{regularization term} in \cref{eq:elbo} is scaled by a hyperparameter $\beta$. 

The choice of $\beta$ creates a trade-off between the reconstruction quality and disentanglement of the latent representation. Previous works have established that increasing the contribution of the \emph{regularization term}, i.e., setting $\beta > 1$, not only promotes independence among latent dimensions but also facilitates the learning of interpretable generative factors (please refer to Appendix~\ref{app:beta_vae} for this literature review). On the other hand, a downside to increasing regularization is the loss of information. When the variational approximation approaches the prior according to the KL term, all encodings $q_\phi(\vec{z}|\vec{x})$ begin to collapse to the prior, resulting in a lack of distinct information about individual data points, which hampers accurate reconstruction. This indicates that setting the value of $\beta$ is non-trivial due to the precarious balance between desirable disentanglement and undesirable loss of information. Even with a suitable parameter $\beta$ for disentangled representation, reconstructed samples may still be blurred due to the loss of information.

\color{black}

%% file: proposed_sec/3_conditionalvae.tex
\setcounter{section}{2}

\begin{figure}
    \centering
    \includegraphics[width=1.0\linewidth]{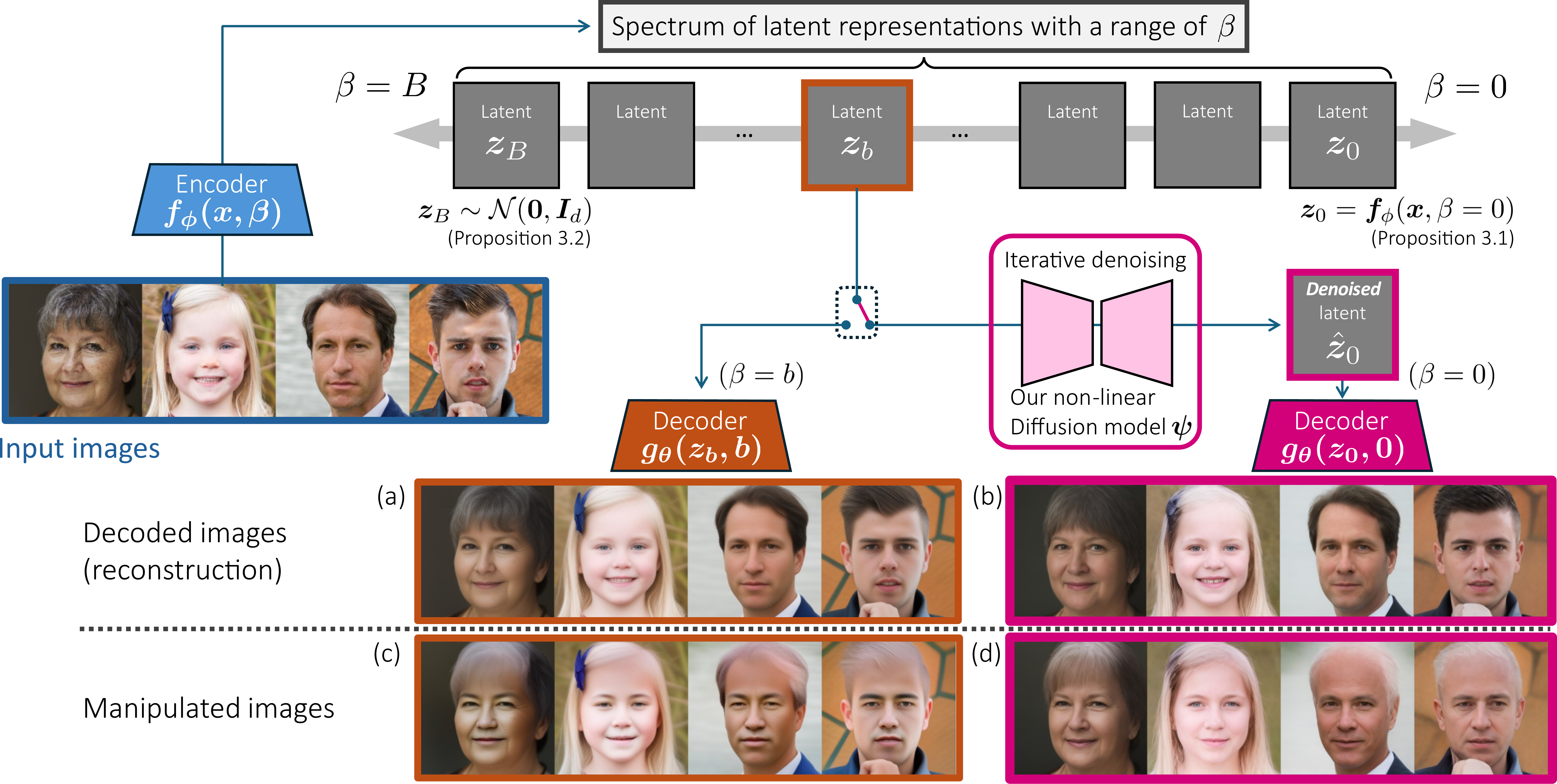}
    \caption{\textbf{Our framework for achieving both disentanglement and generation.} Our approach embeds $\beta$ values as time conditioning in our newly designed nonlinear diffusion model enabling both effective disentanglement and high-quality generation. (a-b) Directly decoding $z_b$ at non-zero $b$ results in blurred images; however, applying the denoiser before decoding yields clear images. (c-d) The denoiser also improves the quality of manipulated images. By using the same direction in the latent space for these manipulations, we achieve consistent changes in age across different ground truth images. This demonstrates that our model produces a disentangled and easily controllable latent space. {\color{c_yuhta}More examples of attribute changes can be found in \cref{app:more_edits}.}}
    \label{fig:architecture}
    \vspace{-10pt}
\end{figure}

In the following sections, we propose our solutions to two critical issues:
\begin{enumerate}[leftmargin=20pt]
    \item \textbf{Problem 1:} Choosing the optimal regularization coefficient ($\beta$) is nontrivial and requires multiple training runs. Our solution to this issue is detailed in \cref{sec:multibeta_rl}.\label{prob:multi_beta}
    \item \textbf{Problem 2:} {\color{c_yuhta}Achieving both generation quality and controllability is challenging, as the regularization used for learning disentangled representations often degrades reconstruction and sample quality. Our solution to this issue is detailed in \cref{sec:latent_diffusion}}\label{prob:nonlin_ldm} 
\end{enumerate}
\section{Multi-$\beta$ Representation Learning}
\label{sec:multibeta_rl}

To overcome the severe trade-off between reconstruction accuracy and the disentanglement of latent representation in existing VAE variants, we propose multi-$\beta$ latent representation learning. First, we extend $\beta$-VAE by treating $\beta$ as a variable rather than a hyperparameter in Section~\ref{subsec:multibeta-VAE}. Using a monotonic property of multi-$\beta$ latent space presented in Section~\ref{subsec:info_loss}, a subsequently learned diffusion model allows us to move across latent spaces corresponding to different $\beta$ (see Section~\ref{sec:latent_diffusion}).

\subsection{Conditional Multi-level $\beta$-VAE}
\label{subsec:multibeta-VAE}

Here we extend the $\beta$-VAE to incorporate $\beta$ as a variable within a range of values. In our setup, unlike the typical $\beta$-VAE, $\beta$ lies in $[0,B]$ instead of being fixed, and it scales the reconstruction and regularization terms with weights of $(B - \beta)$ and $\beta$, respectively (see \cref{eq:condbetaelbo}). This approach allows us to achieve {\color{c_yuhta} a full spectrum of the weighting, encompassing both the reconstruction-only and regularization-only objectives as extreme cases with $\beta=0$ and $B$, respectively.}
We expect that larger values of $\beta$ result in more disentangled latent representations, while smaller values will yield higher fidelity in the reconstructed samples. We assume that each value of $\beta$ has its latent space, which is denoted as $Z_\beta$.
\yuhta{In our VAE, the decoder and encoder for a given $\beta$ are designed as follows:}
\begin{align}
    p_\theta(\vec{x}|\vec{z}_\beta;\beta) &= \mathcal{N}(\vec{x}|\vg_\theta(\vec{z}_\beta,\beta),s^2_\beta \mI)
    \label{eq:cond_generator}\\
    q_\phi(\vec{z}_\beta|\vec{x};\beta) &= \mathcal{N}(\vec{z}_\beta|\vf_\phi(\vec{x},\beta), \sigma^2_\beta \mI),\label{eq:cond_variational}
\end{align}
where $s^2_\beta\in\mathbb{R}_{\geq0},~\sigma^2_\beta\in\mathbb{R}_{\geq0}$, $\theta$ and $\phi$ represent the parameters for the decoder and encoder, respectively. Both the encoder and the decoder depend on $\beta$, which induces different latent spaces. Additionally, the conditional covariance matrices in both the data and latent spaces are modeled as learnable isotropic matrices that depend solely on $\beta$.

Under this model setup, we propose a novel objective function based on a rescaled ELBO as $\mathcal{L}=\mathbb{E}_{\beta}\mathcal{L}_\beta$, where
\begin{align}
    \mathcal{L}_\beta&=\mathbb{E}_{p_{\mathcal{D}}(\vx)}\Bigl[(B - \beta)\mathbb{E}_{q_\phi(\vz_\beta|\vx)}\left[\log p_\theta(\vx|\vz_\beta)\right]
    -\beta \KL(q_\phi(\vz_\beta|\vx)||p(\vz))\Bigr],\label{eq:condbetaelbo}
\end{align}
\yuhta{and we sample $\beta$ from a prior distribution to train the VAE across multiple $\beta$ values.} 
Notably, $\mathcal{L}_\beta$ when $B=1$, with $\beta=0$ and $0.5$, corresponds to the objective functions for a plain autoencoder and a VAE, respectively, without considering the scaling factors\footnote{
\yuhta{Our $\beta$ is different from that of the typical $\beta$-VAE in the relationship between $\beta$ values and their respective models.}}. \cref{algo:ae_training} contains the training algorithm for this {\color{c_yuhta}augmented} $\beta$-VAE.

\subsection{Controlling Information Loss with $\beta$}
\label{subsec:info_loss}

As outlined in the previous section, our objective function~\eqref{eq:condbetaelbo} facilitates a smooth interpolation and extrapolation between the objectives of an autoencoder and a VAE. Specifically, the parameter $\beta$ modulates the degree of information retention in the latent spaces. To elucidate this mechanism, we present a simplified analysis of our novel objective function~\eqref{eq:condbetaelbo} in the case of $B=1$. We anticipate that setting $\beta > 0.5$ encourages disentanglement within the latent space $\mathcal{Z}_\beta$, albeit at the cost of information essential for accurate reconstruction. Conversely, choosing $\beta < 0.5$ enhances reconstruction fidelity but compromises the disentanglement of representations.

A smaller $\beta$ value places greater emphasis on the reconstruction term, resulting in a reduced latent variance $\sigma_\beta^2$.
In the extreme case, setting $\beta=0$ theoretically results in perfect reconstruction with $\sigma_0^2=0$, as demonstrated in the following proposition. We defer the proofs to  Appendix~\ref{app:proofs}.
\begin{proposition}
Under certain regularity conditions, the global optimum of $\mathcal{L}_{0}$ is achieved when $\sigma_0^2=0$.
\label{prop:smallbeta}
\end{proposition}

In contrast, increasing $\beta$ towards 1 enhances regularization, which encourages disentanglement of the latent representation. In the extreme case, when $\beta=1$, the objective function~\eqref{eq:condbetaelbo} reduces to the KL regularization term, causing $q_\phi(\vec{z}|\vec{x};\beta=1)$ collapse to the prior distribution $p(\vz)$. Consequently, the latent space, i.e., $\mathcal{Z}_1$, no longer retains any information about the input $\vx$, as demonstrated in the following proposition.
\begin{proposition}
The mutual information between the input and the reconstructed samples produced by the VAE becomes zero as $\KL(q_\phi(\vec{z}|\vec{x};\beta=1)\parallel p(\vz))$ converges to zero. It holds for any decoder function $g_{\theta}(\cdot,\beta=1)$.
\label{prop:largebeta}
\end{proposition}

The gradual loss of information in the latent space with $\beta$ is also characterised by a gradual increase in the variance of latent representations such that $\sigma_{\beta}<\sigma_{\beta'}$ for $0\leq \beta<\beta'\leq 1$, which is also observed in previous studies~\citep{takida2022preventing}.

In summary, our multi-level $\beta$-VAE is equipped to learn a slew of latent representations which due to \cref{eq:condbetaelbo}, capture major axes of variation present across the dataset by down-weighting accurate reconstructions. This does not mitigate Problem~\ref{prob:nonlin_ldm}.
To this end, we purposely combine the model \cref{eq:cond_variational} and \cref{eq:condbetaelbo} so that learnt $\sigma_\beta$ parallel a typical noising process in diffusion models ($\beta \equiv t$). When trained well, diffusion models can capture the target distribution and sample realisitic data by repeated denoising.
However, due to the involvement of an encoder that specifies the mean $\vf_\phi(\vec{x},\beta)$ at all $\beta \in [0,1]$ in the latent space, our noising process diverges from the commonly used linear inference/noising process. We detail our formulation of non-linear denoising diffusion in the next section.

\color{black}

%% file: proposed_sec/3_ldm.tex
\section{Reversing the Information Loss}
\label{sec:latent_diffusion}

Increasing regularization enhances representation learning but negatively affects sample and reconstruction quality. At higher $\beta$ values, the reconstructions tend to collapse into an “averaged” image, a phenomenon also noted by~\citet{Collins:2022qpr}. To address Problem 2, we propose reversing information loss by training a denoising model based on a diffusion process. First, we review the standard diffusion model in \cref{subsec:primer_lindiffusion} and its nonlinear extension in \cref{subsec:non_linear_dm}. In this approach, the time-varying mean is governed by the encoder ($\vf_\phi$), with noise conditioning parameterized by $\beta$ or equivalently by time $t$.

\subsection{Primer on Diffusion Models}
\label{subsec:primer_lindiffusion}
We start with a brief primer on the vanilla diffusion models with a linear diffusion process.
Diffusion models consist of a fixed hierarchical encoding process, known as the forward or noising process, and a decoding process for generation. In the encoding stage, incremental noise is gradually added to the data, transforming it into a Gaussian noise:
\begin{equation}
    q(\vec{z}_t | \vec{x};t ) := \mathcal{N}(\vec{z}_t | \vec{x},\sigma^2_t\mI_d),\label{eq:linear_marginal}
\end{equation}
where $t \in [0,T]$, $\tau>0$ is a small constant, and $\sigma_t >0$ is a predefined noise schedule that increases with $t$. Next, the Markovian forward distributions are derived as
\begin{align}
    &q(\vz_t|\vz_{t-\tau}) = \mathcal{N}(\vz_t|\vz_{t-\tau},\sigma^2_{t|t-\tau}\mI_d), \label{eq:linear_forward}%
\end{align}
where $\sigma^2_{t|t-\tau} = \sigma^2_t - \sigma^2_{t-\tau}$.
A tractable reverse decoding process is obtained via Bayes' rule as
\begin{align}
    &q(\vz_{t-\tau}|\vz_t,\vx) = \mathcal{N}(\vz_{t-\tau}|\yuhta{\tilde{\vec{\mu}}_t(\vz_t,\vx)},\yuhta{\tilde{\sigma}_t^2}\mI_d), \label{eq:posterior_forward}\\
    &\text{where }\yuhta{\tilde{\sigma}_t^2} = \frac{\sigma^2_{t|t-\tau}\sigma^2_{t-\tau}}{\sigma^2_t} \text{ and } \,\,\,\yuhta{\tilde{\vec{\mu}}_t(\vz_t,\vx)} = \frac{\sigma^2_{t-\tau}}{\sigma^2_t} \vz_t + \frac{\sigma^2_{t|t-\tau}}{\sigma^2_t}\vx.\label{eq:mu_sigma_linear_posterior}
\end{align}
Diffusion models are trained to match the generative reverse conditional distributions in \cref{eq:posterior_forward}, and are generally parameterized as
\begin{align}
    &p_\psi(\vz_{t-\tau}|\vz_t) := \mathcal{N}(\vz_{t-\tau}|\yuhta{\vec{\mu}_\psi(\vz_t,t)},\sigma^2_r(t)\mI_d),\label{eq:generative_cond} \\ 
    &\text{where }\yuhta{\vec{\mu}_\psi(\vz_t,t)} := \frac{\sigma^2_{t-\tau}}{\sigma^2_t} \vz_t + \frac{\sigma^2_{t|t-\tau}}{\sigma^2_t}\hat{\vx}_\psi(\vz_t,t)\label{eq:lin_model_mean}.
\end{align}
$\hat{\vx}_\psi$ is  known as the \emph{denoiser}. Equivalently, we parametrize it as a noise prediction model $\hat{\bm{\epsilon}}_\psi$, where
$\hat{\vx}_\psi(\vz_t, t)=\vz_t - \sigma_t \hat{\bm{\epsilon}}_\psi(\vz_t, t)$. %
The loss function of a diffusion model is derived using an ELBO (by extending \cref{eq:elbo} with time hierarchy), aiming to match the conditional distributions in \cref{eq:posterior_forward} and \cref{eq:generative_cond} over all $t\in [0,T]$ using the KL divergence. This loss boils down to a simple regression loss:
\begin{equation}
    \min_\psi\yuhta{\mathbb{E}_{t,\vx}\mathbb{E}_{\vz_t|\vx}}\left[\frac{1}{2\yuhta{\tilde{\sigma}_t^2}}\|\yuhta{\vec{\mu}_\psi(\vz_t,t)} - \yuhta{\tilde{\vec{\mu}}_t(\vz_t,\vx)}\|^2_2\right].\label{eq:clean_pred}
\end{equation}
By discretizing the time such that $t\in\{iT/N\}_{i=0}^N$ for the iterative decoding, we obtain a hierarchical generator: 
\begin{equation}
    p_\psi(\vec{x}) = \int_{\vz} p(\vec{x}|\vec{z}_0)p(\vec{z}_B)\prod_{i=1}^{\yuhta{N}} p_\psi\left(\vec{z}_{\yuhta{\frac{(i-1)T}{N}}}\Big\vert\vec{z}_{\yuhta{\frac{iT}{N}}}\right),
    \label{eq:diffusion_gen_model}
\end{equation}
where $p(\vec{z}_T) = \mathcal{N}(\vec{z}_T|0,\mI_d)$.

\subsection{Non-linear Diffusion in Latent Space}\label{subsec:non_linear_dm}

We propose a non-linear (in $\vec{x}$) denoising diffusion for use in our model. In this subsection, time variables $t$ (and $T$) are interchangeable with $\beta$ (and $B$), as they represent the same concept. We hinted in \cref{subsec:info_loss} that our non-linear diffusion process is prescribed by the $\beta$- or time-dependent encoder. Formally, the distribution of $\vz_t$ for given $\vx$ is
\begin{equation}
     q_\phi(\vec{z}_t|\vec{x};t) = \mathcal{N}(\vec{z}_t|{\color{blue}\vf_\phi(\vec{x},t)}, \sigma^2_t \mI_d).
     \label{eq:nonlin_marginal}
\end{equation}
This expression is just an adaptation of \cref{eq:cond_variational} with $\beta=t$, and is more general than \cref{eq:linear_marginal}. We propose the nonlinear Markovian encoding process as
\begin{equation}
    q_\phi(\vz_t|\vz_{t-\tau},\vx) = \mathcal{N}(\vec{z}_t|\vz_{t-\tau} + {\color{blue}\vf_\phi(\vec{x},t) - \vf_\phi(\vec{x},t-\tau)},\sigma^2_{t|t-\tau}).\label{eq:nonlin_forward}
\end{equation}
This form closely resembles \cref{eq:linear_forward}. Following the development in \cref{subsec:primer_lindiffusion}, we now define the reverse of \cref{eq:nonlin_forward} as:
\begin{align}
    &q_\phi(\vz_{t-\tau}|\vz_t,\vx) = \mathcal{N}(\vec{z}_{t-\tau}|\tilde{\vec{\mu}}_t(\vz_t,\vx),\tilde{\sigma}_t^2\mI_d),\label{eq:nonlin_posterior_forward}\\
    &\text{where }\tilde{\vec{\mu}}_t(\vz_t,\vx) = \frac{\sigma^2_{t-\tau}}{\sigma^2_t} \vz_t + \frac{\sigma^2_{t|t-\tau}}{\sigma^2_t}\vx+ {\color{blue}\vf_\phi(\vec{x},t-\tau) - \vf_\phi(\vec{x},t)},\label{eq:non_lin_mean}
\end{align}
with $\tilde{\sigma}_t^2$ remaining the same as \cref{eq:mu_sigma_linear_posterior}.  
Due to the additional $\vf_\phi$ terms in this flavour of the diffusion model, $\vec{\mu}_\psi$ cannot follow the same parameterization as in \cref{eq:lin_model_mean}. Instead we introduce a new approach to express the mean prediction in this case, as follows:
\begin{equation}
    \yuhta{\vec{\mu}_\psi(\vz_t,t)} := \frac{\sigma^2_{t-\tau}}{\sigma^2_t} \vz_t + \frac{\sigma^2_{t|t-\tau}}{\sigma^2_t}\hat{\vx}_\psi(\vz_t,t) +{\color{blue}\hat{\vec{\Delta}}_\psi(\vz_t,t)}.\label{eq:nonlin_model_mean}
\end{equation}

\yuhta{\cref{eq:nonlin_model_mean}} introduces an extra predictor $\hat{\vec{\Delta}}_\psi$ for learning the evolution of encodings with time. In practice, we train noise prediction, reparameterizing $\hat{\vx}_\psi(\vz_t,t)$ with $\hat{\vec{\epsilon}}_\psi(\vz_t,t)$, along with an encoding difference predictor $\hat{\vec{\Delta}}_\psi(\vz_t,t)$, which is novel to the best of our knowledge.

\begin{figure}[t]
    \centering

\begin{tabular}{p{0.47\textwidth}p{0.47\textwidth}}
\begin{algorithm}[H]
\small
\caption{Training of our VAE}
\label{algo:ae_training}
\DontPrintSemicolon
\KwIn{\yuhta{Dataset $\mathcal{D}$}, \yuhta{$\beta$-schedule $\{\beta_i\}_{i=1}^N$}, learning rate $\eta$, number of training steps $S$}

\KwOut{Trained networks, $\phi$, $\theta$, and \yuhta{$\bm{\sigma}=\{\sigma_\beta\}_{\beta\in\{\beta_i\}_{i=1}^N}$}}

\BlankLine

\For{$s = 1, 2, \dots, S$}{
\Indp
 \nl Sample: $\vx \sim p_{\mathcal{D}}(\vx),\ \beta \sim \mathcal{U}([0,1]), \  \bm{\epsilon} \sim \mathcal{N}(\mathbf{0}, \mI_d)$\;
 \BlankLine
 \nl Generate noisy encoding: $\vz_\beta = \vf_\phi(\vx,\beta) + \sigma_\beta \bm{\epsilon}$\;
\BlankLine
\nl Compute the objective $\mathcal{L}$ based on \cref{eq:condbetaelbo}\;
\BlankLine 
\nl Update parameters: $\omega \leftarrow \omega - \eta \nabla_\omega \mathcal{L}$, where $\omega= \{\theta, \phi, \bm{\sigma}\}$ 
}
\Return $\vf_\phi, \vg_\theta, \sigma$\;
\end{algorithm}

&
\begin{algorithm}[H]
\small
\caption{Sampling of our diffusion}
\label{algo:sampling}
\DontPrintSemicolon

\KwIn{Trained model $\vec{\epsilon}_\psi$, total time steps $N$, largest time $T$, trained noise schedule $\vec{\sigma}=\{\sigma_t\}_{t\in \{0,T/N,\ldots,(N-1)T/N,T\}}$}
\KwOut{Generated sample $\vz_0$}
\SetNlSkip{0em}
\SetNlSty{}{}{.}
\BlankLine
\textbf{Initialize:} $\vz_T\sim \mathcal{N}(\mathbf{0},\mI_d)$ 
\BlankLine
\For{$t = T, (N-1)T/N \ldots, T/N$}{
\Indp
    
    \nl Predict noise and diff: \;
     $\hat{\vec{\epsilon}} = \hat{\vec{\epsilon}}_\psi(\vz_t, t),\;\hat{\vec{\Delta}}=\hat{\vec{\Delta}}_\psi(\vz_t,t)
     $\;
     \BlankLine
    \nl Compute mean for $z_{t}$:\;
    $
    \vec{\mu}_t =  \vz_t -  \sigma_t\hat{\epsilon}
    $
    \BlankLine
    \nl Predict previous mean: %
    $
    \vec{\mu}_{t-T/N} = \vec{\mu}_t - \hat{\vec{\Delta}}
    $
    \BlankLine
    \nl Update $\vz_{t-T/N}$ (with $\vec{\varepsilon} \sim \mathcal{N}(\bm{0},\bm{I}_d)$):%
    $
    \vz_{t-T/N} = \vec{\mu}_{t-T/N} + \sigma_{t-T/N} \vec{\varepsilon}
    $
}
\Return $\vz_0$\;

\end{algorithm}

\vspace{-70pt}
\end{tabular}
\end{figure}

Following the parameterization of $\vec{\mu}_\psi$, our model training differs from standard practice in a couple of key ways. First, we do not train the noise prediction network to predict the noise added to $\vz_0=f_\phi(\vx,0)$ given a sample $\vz_{t}$. The transition from $\vz_0$ to $\vz_t$ is non-linear due to the encoder and depends on its Jacobian, $\frac{\dif \vf_\phi(\vec{x},t)}{\dif t}$. Rather than learning an inversion of this time-varying encoding, we aim to learn the direction of the noise ($\vec{\epsilon}_{t}$) at each time step $t\in[0,T]$ using $\hat{\vec{\epsilon}}_\psi(\vz_t,t)$, i.e.,
\cyuhta
\begin{align}    
\mathbb{E}_{t,\vx}\mathbb{E}_{\vec{\epsilon}_t}\mathbb{E}_{\vz_t|\vx,\vec{\epsilon}_t}\left[\frac{1}{w(t)}\|\hat{\vec{\epsilon}}_\psi(\vz_t,t) - \vec{\epsilon}_t \|^2_2\right]\text{ with }
\vz_{t} = \vf_\phi(\vec{x},t) + \sigma_t\vec{\epsilon}_{t}\text{ and } \vec{\epsilon}_t\sim\mathcal{N}(\mathbf{0},\mI_d),
\end{align}
\color{black}
where $w(t)$ is \yuhta{a weighting function}. This approach trains the model to denoise at each time step. Additionally, $\hat{\vec{\Delta}}_\psi$ is necessary for sampling and is trained to predict the change in $\vf_\phi$ over a small time interval $\tau${\footnote{In implementation, we set $\tau=T/N$ in the discrete time setup, learning the single time-step encoding difference for all times.}}. Assuming that the encoder is a smooth function, 
this design is based on the intuition that learning encoding differences over one time step is easier than over arbitrarily large steps.

We adjust the diffusion model’s U-Net to produce two outputs, $\hat{\vec{\epsilon}}(\vz_t, t),\:\hat{\vec{\Delta}}_\psi(\vz_t,t)$. Using these predictions, we build our DDPM~\citep{hoddpm20}-inspired sampling algorithm, as shown in \cref{algo:sampling}. The actual loss function used to train this new diffusion model is defined as
\begin{align}
\mathbb{E}_{t,\vx}\mathbb{E}_{\vec{\epsilon}_t}\mathbb{E}_{\vz_t|\vx,\vec{\epsilon}_t}\Bigl[\frac{1}{w(t)}\|\hat{\vec{\epsilon}}_\psi(\vz_t,t) - \vec{\epsilon}_t(\vx,\vec{\epsilon}_t) \|^2_2 +{\color{blue}\|\vf_\phi(\vec{x},t-\tau) - \vf_\phi(\vec{x},t) - \hat{\vec{\Delta}}_\psi(\vz_t,t)\|_2^2}\Bigr].   
\end{align}
Integrating the conditional multilevel $\beta$-VAE, as introduced in \cref{subsec:multibeta-VAE}, with the diffusion model described in \cref{subsec:non_linear_dm} is key to mitigating the disentanglement-reconstruction trade-off in our framework.

We now combine the two proposed modules from \cref{sec:multibeta_rl} and \cref{sec:latent_diffusion} into a single model, trained in two phases. In the first phase, we train the conditional multilevel $\beta$-VAE using the loss defined in \cref{eq:condbetaelbo} (\cref{algo:ae_training}). After this, we train the non-linear diffusion model from \cref{subsec:non_linear_dm}, keeping the autoencoder parameters fixed. Additionally, depending on the dataset, we fine-tune the decoder with an adversarial loss, following~\citet{Rombach2021HighResolutionIS}, to enhance generation quality. Further training details are provided in \cref{app:exp_deets}.

%% file: proposed_sec/4_relatedworks.tex
\setcounter{section}{4}
\section{Related Works}
\label{sec:related_works}

$\beta$-VAE and its variants have been extensively studied for their distinct capabilities~\citep{burgess2018understandingdisentanglingbetavae} and wide-ranging applications in domains such as images~\citep{Higgins2016betaVAE}, text~\citep{controlVAEshao20b}, and molecular generation~\citep{richards2022conditional}. These models are especially valued for their interpretable latent representations, achieved through $\beta$-controlled regularization.

\citet{Kim2018DisentanglingBF} improved disentanglement while maintaining reconstruction quality by combining the ELBO with a total correlation term. Similarly, \citet{chen2018isolating} enhanced mutual information between latent variables and observed data to promote independence among latent factors. \citet{controlVAEshao20b} introduced an adaptive feedback mechanism that adjusts $\beta$ during training based on KL divergence. \citet{Dewangandeepconv22} applied a deep convolutional $\beta$-VAE for feature extraction in industrial fault diagnosis, using a variable $\beta$ training protocol without conditioning the encoder and decoder on $\beta$.

Both \citet{Collins:2022qpr} and \citet{bae2023multirate} investigated training VAEs with multiple $\beta$ values, with \citet{bae2023multirate} capturing the full rate-distortion curve using a hypernetwork conditioned on $\beta$, and \citet{Collins:2022qpr} focusing on particle physics applications. In contrast, our framework enables high-fidelity generation from disentangled representations (with larger $\beta$) by linking latent spaces through a novel non-linear diffusion process, along with specific adjustments to the $\beta$-conditioned VAE to optimize the latent space for diffusion.

Recently, a few methods have emerged to achieve disentanglement while preserving generation fidelity. \citet{wang23Info} developed InfoDiffusion, a pioneering diffusion-based model that extends the Diffusion Autoencoder~\citep[DiffAE,][]{preechakul2022diffusion}. \citet{yang2023disdiff} introduced DisDiff, which adds encoder and decoder components to pre-trained diffusion models to maintain generation quality, rather than training disentanglement-focused models from scratch. \citet{ren2022DisCo} focus on building an exploration technique for pre-trained generative models for post-hoc identification of disentangled directions. Notably, in contrast to training-free methods our approach encourages disentanglement during training, assuring a more disentangled representation.

Our method falls within the model-training-based category, similar to InfoDiffusion, but it uniquely learns a spectrum of latent representations that provide a distinct advantage. Combined with the learned non-linear diffusion model, our framework enables transitions between highly disentangled and fully informed latent representations, facilitating the generation of high-fidelity outputs.Additionally, we show in \cref{app:ours_vs_disco} (see \cref{table:moreTab1}) that finding-based methods like DisCo \citep{ren2022DisCo} can complement our approach, indicating potential for combined implementations.

%% file: sec/5_experiments.tex
\section{Experiments}
\label{sec:experiments}
We quantitatively demonstrate that our method achieves both disentanglement and high-quality generation within a single model. We will focus on each of these aspects individually. Specifically, in \cref{subsec:exp1}, we examine the disentanglement of the learned latent representations. In \cref{subsec:exp2}, we evaluate our model’s unconditional generation performance and report standard metrics on commonly used image datasets.  Additional details regarding the experiments can be found in the appendix.

\begin{wraptable}{r}{0.45\textwidth}
\centering
\small
\vspace*{-45pt}
\caption{TAD and FID scores on CelebA. Our model outperforms baselines in terms both of disentanglement and generation quality.}
\label{table:tradeoff_results}
\small
\begin{tabular}{lcc}
\toprule
Method & TAD ($\uparrow$) &  FID ($\downarrow$) \\
\midrule
$\beta$-VAE         & $0.088 \pm 0.043$ & $99.8 \pm 2.4$ \\
InfoVAE             & $0.000 \pm 0.000$ & $77.8 \pm 1.6$ \\
DiffAE     & $0.155 \pm 0.010$ & $22.7 \pm 2.1$ \\
InfoDiffusion       & $0.299 \pm 0.006$ & $23.6 \pm 1.3$ \\
DisDiff & $0.305 \pm 0.010$ & $18.3 \pm 2.1$ \\
Ours & $\mathbf{0.378 \pm 0.017}$ & $\mathbf{17.9 \pm 1.9}$\\
\bottomrule
\end{tabular}
\vspace*{-10pt}
\end{wraptable}

\subsection{Evaluating Disentanglement}
\label{subsec:exp1}

\subsubsection{Image Dataset}
To demonstrate that our method effectively obtains disentangled representation while maintaining high generation quality, we verify our approach on an image dataset. Specifically, we follow the protocol established by \citet{yang2023disdiff} using the CelebA dataset~\citep{liu2015deep}. In this setup, we calculate Total AUROC Difference~\citep[TAD,][]{yeats2022nashae} and FID~\citep{fid17heusel} scores. TAD is a disentanglement metric for datasets with binary attribute labels that measures how well latent variables uniquely capture ground truth attributes. As shown in Table~\ref{table:tradeoff_results}, our method achieves the best performance in both aspects compared to the baselines, including state-of-the-art methods that aim to address the significant trade-off between these two factors, such as InfoDiffusion~\citep{wang23Info} and DisDiff~\citep{yang2023disdiff}.

\subsubsection{Common Benchmark with Toy Datasets}
\label{subsec:exp1_toydata}
To compare our model with baselines dedicated to disentanglement (at the expense of generation quality) such as FactorVAE~\citep{Kim2018DisentanglingBF}, $\beta$-TCVAE~\citep{chen2018isolating}, and InfoGAN-CR~\citep{lin20crgan}, we adopt the evaluation protocols of~\citet{pmlr-v97-locatello19a} and~\citet{Khrulkov2021DisentangledRF} using toy datasets: Cars3D~\citep{reedcars15}, Shapes3D~\citep{Kim2018DisentanglingBF}, and MPI3D~\citep{Gondal2019OnTT}. For assessment, we use the Mutual Information Gap (MIG)~\citep{chen2018isolating} and the Disentanglement metric~\citep{eastwood2018a}. The MIG measures how well latent dimensions respond to changes in individual generative factors, while DCI (Disentanglement, Completeness, and Informativeness) evaluates the extent to which factors are distinctly represented by individual latent dimensions, the exclusivity of each factor to a specific dimension, and the comprehensiveness of the overall representation. \cref{table:dis_results} shows that our model achieves disentanglement performance comparable to or surpassing that of other disentanglement models.

We also compare our approach with DisCo~\citep{ren2022DisCo}, another approach that uses contrastive learning to find the appropriate directions in the latent space of a pre-trained generative model while preserving its generative capabilities. We demonstrate the finding-based approach is complementary to our method in \cref{app:ours_vs_disco} (see \cref{table:moreTab1}).

\begin{table}[tb]
    \centering
    \scriptsize
    \sisetup{separate-uncertainty=true, table-format=+1.2(3),retain-zero-uncertainty=true, detect-weight,mode=text}
    \caption{\textbf{Disentanglement Metrics.}
    We evaluate our multi-$\beta$ VAE representations through benchmarking on well-known toy datasets and comparing them to baselines that aim to learn disentangled representation. In each column, the best results are highlighted in bold, and the second-best results are underlined.
    }
    \label{table:dis_results}
    \renewrobustcmd{\bfseries}{\fontseries{b}\selectfont}
    \renewrobustcmd{\boldmath}{}
    \newrobustcmd{\B}{\bfseries}
    \begin{tabular}{lcccccc}
         \toprule
         & \multicolumn{2}{c}{Cars3D} & \multicolumn{2}{c}{Shapes3D} & \multicolumn{2}{c}{MPI3D}\\
         \cmidrule(r){2-3}\cmidrule(l){4-5}\cmidrule(l){6-7}
         Method & {MIG$\uparrow$} & {DCI$\uparrow$} & {MIG} & {DCI} & {MIG} & {DCI}\\
         \midrule
         FactorVAE & \textbf{0.128} $\pm$ 0.036 & \textbf{0.160} $\pm$ 0.020 & \underline{0.411} $\pm$ 0.163 & 0.611 $\pm$  0.127 & 0.098 $\pm$ 0.027 & \underline{0.246} $\pm$ 0.066\\
         $\beta$-TCVAE & 0.080 $\pm$ 0.024 & 0.140 $\pm$ 0.020 & 0.406 $\pm$ 0.190 & \underline{0.613} $\pm$ 0.151 & 0.108 $\pm$ 0.053 & 0.239 $\pm$ 0.062\\
         InfoGAN-CR & 0.011 $\pm$ 0.009 & 0.020 $\pm$ 0.011 & 0.297 $\pm$ 0.124 & 0.478 $\pm$ 0.055 & \textbf{0.161} $\pm$ 0.077 & 0.242 $\pm$ 0.076\\
         Ours & \underline{0.117} $\pm$ 0.009 & \underline{0.157} $\pm$ 0.010 & \textbf{0.422} $\pm$ 0.090 & \textbf{0.621} $\pm$ 0.090 & \underline{0.147} $\pm$ 0.035 &  \textbf{0.253} $\pm$ 0.043\\
        \bottomrule
    \end{tabular}
\end{table}

\subsection{Evaluating Generation Quality}
\label{subsec:exp2}
We demonstrate that our model can serve as a standalone generation model by evaluating its generation quality on practical image datasets, including CelebA-HQ~\citep{karras2018progressive}, FFHQ~\citep{karras19styleGAn}, and LSUN-Bedrooms~\citep{fisherLSUN}, at a resolution of $256\times256$. For these image datasets, we evaluate our model using FID to assess image quality and precision-recall~\citep{Kynknniemi2019ImprovedPA} to gauge data distribution coverage. Details of our architecture and training times are provided in \cref{app:exp_deets}, and generated sample images are shown in \cref{app:samples}.

While our model is designed to learn disentangled latent representations, it is crucial that this capability does not compromise generation quality. Based on our comparative performance with generation-focused baselines in \cref{tab:evaluation}, we conclude that our model effectively generates high-quality images at this resolution across various datasets.

\begin{figure}
    \centering
        \includegraphics[width=1.0\linewidth,trim=5 5 20 5,clip]{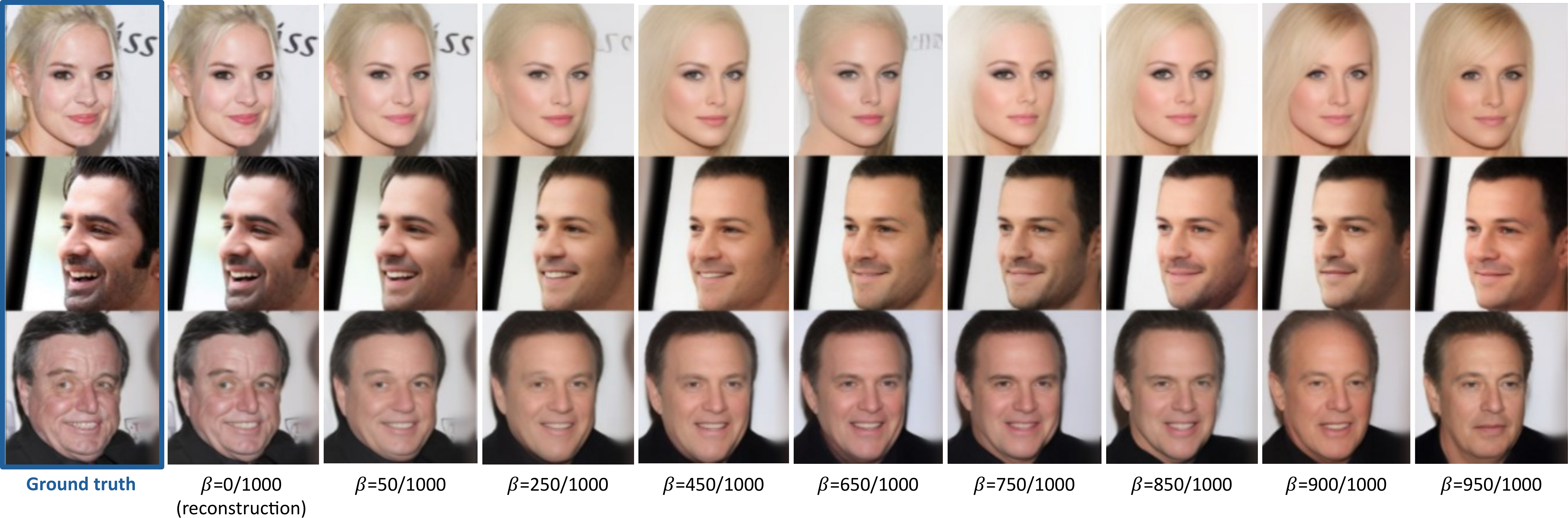}
    \caption{\textbf{Smoothness of latent space in $\beta$ ($t$):}
    We provide evidence for the smoothness of the learned representations by generating samples from latent spaces with various values of $\beta$.
    Notably, latent representations obtained by smaller $\beta$ values tend to produce images closer to ground truth because they retain more information.}
    \label{fig:smooth_latent}
\end{figure}

Lastly, we visualize generated samples from various $\beta$ values in \cref{fig:smooth_latent} to demonstrate the smoothness of our latent space spectrum in terms of $\beta$ or $t$. We encode ground truth images to latent representations at certain $\beta$ values, denoise them using our non-linear diffusion, and decode them to obtain clear images. We observe that generated images with different values of $\beta$ remain consistent with each other. Another smoothness perspective, spatial smoothness of the latent spaces, is visualized in \cref{subsec:exp4}.

\begin{table*}[tb]
\large
\centering
\setlength{\tabcolsep}{1pt}
\caption{\textbf{Generation Quality.}
We evaluate our model for unconditional image synthesis and report standard metrics, comparing them against baselines specifically designed for generation.}
\resizebox{\linewidth}{!}{
\begin{tabular}{@{}lccclccclccc@{}}
\toprule
\multicolumn{4}{c}{\textbf{CelebA-HQ 256 $\times$ 256}} & \multicolumn{4}{c}{\textbf{FFHQ 256 $\times$ 256}} & \multicolumn{4}{c}{\textbf{LSUN-Bedrooms 256 $\times$ 256}} \\ \cmidrule(r){1-4}\cmidrule(l){5-8}\cmidrule(l){9-12}
\textbf{Method} & \textbf{FID $\downarrow$} & \textbf{Prec. $\uparrow$} & \textbf{Recall $\uparrow$} & \textbf{Method} & \textbf{FID $\downarrow$} & \textbf{Prec. $\uparrow$} & \textbf{Recall $\uparrow$}&\textbf{Method} & \textbf{FID $\downarrow$} & \textbf{Prec. $\uparrow$} & \textbf{Recall $\uparrow$}\\ \midrule
DC-VAE\citep{Parmar_2021_CVPR}  & 15.8 & - & - & ImageBART\citep{esser2021imagebart}  & 9.57 & - & - & ImageBART\citep{esser2021imagebart}  & 5.51 & - & - \\
VQGAN+T\citep{Esser_2021_CVPR} & 10.2 & - & - &U-Net GAN\citep{Schonfeld_2020_CVPR}   & 10.9 & - & -&DDPM\citep{hoddpm20} & 4.9 & - & - \\
PGGAN\citep{karras2018progressive}  & 8.0 & - & - &UDM\citep{Kim2021SoftTA}  & 5.54 & - & - &UDM\citep{Kim2021SoftTA} & 4.57 & - & - \\
LSGM\citep{vahdat2021scorebased}  & 7.22 & - & -&StyleGAN\citep{karras19styleGAn}  & 4.16 & 0.71 & 0.46&StyleGAN\citep{karras19styleGAn} & 2.35 & 0.59 & 0.48\\
UDM\citep{Kim2021SoftTA}  & 7.16 & - & -& ProjectedGAN\citep{sauer2021projectedgans}  & 3.08 & 0.65 & 0.46&ADM\citep{dhariwal2021diffusion} & 1.90 & 0.66 & 0.51 \\ \midrule
Ours & 6.41 & 0.71 & 0.48 & Ours & 5.45 & 0.72 & 0.48 &Ours & 3.2 & 0.65 & 0.48\\
\bottomrule

\end{tabular}}
\label{tab:evaluation}
\end{table*}

%% file: sec/6_futurework.tex
\section{Conclusion}
We propose a new generative modeling framework that leverages a range of $\beta$ values to learn disentangled representations and sharp generation quality, including unconditional generation. Our framework introduces two key components: (1) a multi-$\beta$ VAE, producing a spectrum of latent representations that can be refined via a denoising diffusion process, and (2) a non-linear diffusion model that links latent representations for different $\beta$ values.
Our method offers a superior trade-off compared to existing approaches. Additionally, it achieves comparable disentanglement performance to dedicated baselines while maintaining high decoding quality and generating results on par with state-of-the-art generation models.

\color{black}

%% file: app/beta_vae.tex
\section{Supplementary for $\beta$-VAE}
\label{app:beta_vae}

Disentanglement induced by larger $\beta$ in $\beta$-VAE can be explained by breaking down the regularization term as outlined by \citet{hoffman2016elbo,Makhzani2015AAE}: 
\begin{equation}
    \mathbb{E}_{p_{\mathcal{D}}(\vec{x})}\left[\KL(q_\phi(\vec{z}|\vec{x}) || p(\vec{z}))\right] = \mathcal{I}(\vec{z};\vec{x}) + \KL(q_\phi(\vec{z})||p(\vec{z}))
\end{equation}
Here, $\mathcal{I}(\vec{z};\vec{x})$ represents the mutual information between the latent variables and data points, and $q_\phi(\vec{z})$ is the marginal posterior distribution. Setting $\beta > 1$ forces $q_\phi(\vec{z})$ to be more similar to the factorized prior distribution, promoting independence within the latents. This also reduces the mutual information between the data and the latent variables, leading to information loss and poor reconstructions.

While the breakdown above explains why latent factors learned with $\beta > 1$ are more independent, it does not clarify why increasing the contribution of the \emph{regularization term} leads to learning interpretable generative factors. In a dataset consisting of a variety of three-dimensional objects, these generative factors would include object type, orientation, scale, lighting, and color. Here, we subscribe to the hypothesis presented by \citet{burgess2018understandingdisentanglingbetavae, Mathieu2018DisentanglingDI}, which posits that by forcing the variational posterior $q_\phi(\vec{z}|\vec{x})$ to be more similar to the prior $p(\vec{z})$, we can achieve two desirable outcomes: (1) the scale of the prior forces the representations across data points to overlap, which inherently causes prominent generative attributes to overlap in the latent space; (2) the prior term also forces the variational posterior to be more factorized across data points, which results in different axes in $\vec{z}$ corresponding to distinct generative factors.

%% file: app/proofs.tex
\section{Proofs}
\label{app:proofs}

\subsection{Proposition~\ref{prop:smallbeta}}
Formal restatement of Proposition~\ref{prop:smallbeta} is provided as Proposition~\ref{prop:smallbeta_formal} below.

\begin{proposition}
    \textit{Assume that $p_{\mathcal{D}}(\vx)$ has finite covariance and $f_{\phi}$ is arbitrarily complex, provided it is Lipschitz continuous. For the global optimum of $\mathcal{L}_\beta$ with respect to $\sigma_\beta^2$, we have $\sigma_\beta^2 \to 0$ as $\beta \to 0$.}
    \label{prop:smallbeta_formal}
\end{proposition}
The two assumptions in this proposition are reasonable in practice. Since $p_{\mathcal{D}}(\vx)$ is expressed as a sum of Dirac delta functions $\delta(\vx-\vx_i)$ for $i = {1, \ldots, M}$, where each $\vx_i$ represents a set of bounded pixel values, its covariance is finite. Furthermore, because $f_{\phi}$ is implemented within a deep neural network framework, it is expected to have strong representational capacity and to be Lipschitz continuous~\citep{fazlyab2019efficient}. This proposition follows almost directly from Theorem~3 in \citet{takida2022preventing}. For completeness, we provide the proof in this manuscript.
\begin{proof}
For a given $\beta$, consider the input sample $\vx$ and the reconstructed sample $\vx^\prime = g_{\theta}(\vz_\beta, \beta)$ obtained via the stochastic latent variable $\vz_\beta \sim q_{\phi}(\vz_\beta|\vx; \beta)$. Let $\vz_\beta^e$ denote the encoded latent variable without applying the reparameterization trick, i.e., $\vz_\beta^e = f_{\phi}(\vx, \beta)$. For improved readability, we omit $\beta$ in probability distributions (e.g., $q_{\phi}(\vz_\beta|\vx; \beta)$). Define the marginalized distributions for $\vz_\beta^e$ and $\vz_\beta$ as $q_{\phi}(\vz_\beta^e) := \mathbb{E}_{p_\mathcal{D}(\vx)}[\delta(\vz_\beta^e - f_{\phi}(\vx, \beta))]$ and $q_{\phi, \sigma_\beta^2}(\vz_\beta) := \mathbb{E}_{p_\mathcal{D}(\vx)}[q_{\phi}(\vz_\beta|\vx)]$, with their respective covariance matrices denoted as $\vec{\Sigma}_{\phi}^e$ and $\vec{\Sigma}_{\phi, \sigma_\beta^2}$. Additionally, let $q_{\sigma_\beta^2}(\vz_\beta|\vz_\beta^e)$ represent the conditional distribution of $\vz_\beta$ given $\vz_\beta^e$.

By applying the data processing inequality to the Markov chain $\vx \to \vz_\beta^e \to \vz_\beta \to \vx^\prime$, the following holds: \begin{align} 
    \mathcal{I}(\vz_\beta, \vz_\beta^e) \geq \mathcal{I}(\vx, \vx^\prime).
\label{eq:ineq1}
\end{align}
Furthermore, the mutual information between $\vz_\beta$ and $\vz_\beta^e$ is bounded above as follows:
\begin{align}
    \mathcal{I}(\vz_\beta,\vz_\beta^e)
    & = \mathcal{H}[q_{\phi,\sigma_\beta^2}(\vz_\beta)]-\mathbb{E}_{q_{\phi}(\vz_\beta^e)}\mathcal{H}[q_{\sigma_\beta^2}(\vz_\beta|\vz_\beta^e)] \\
    &\leq H(\vec{\Sigma}_{\phi,\sigma_\beta^2})-H(\sigma_\beta^2\vec{I})\\
    &=\frac{d}{2}\log\left(\frac{\mathrm{det}(\vec{\Sigma}_{\phi,\sigma_\beta^2})}{\sigma_\beta^2}\right),
    \label{eq:ineq2}
\end{align}
where $\mathcal{H}(p)$ and $H(\vec{\Sigma})$ denote the differential entropy of the distribution $p$ and the Gaussian distribution with covariance $\vec{\Sigma}$, respectively.
By combining Equations~\eqref{eq:ineq1} and \eqref{eq:ineq2}, we arrive at the following: 
\begin{align}
    \mathcal{I}(\vx,\vx^\prime)\leq\frac{d}{2}\log\left(\frac{\mathrm{det}(\vec{\Sigma}_{\phi,\sigma_\beta^2})}{\sigma_\beta^2}\right).
    \label{eq:ineq_main}
\end{align}

To prove that $\sigma_\beta^2\to0$ as $\beta\to0$, we show that (i) $\mathcal{I}(\vx,\vx^\prime)$, the LHS of \Cref{eq:ineq_main}, diverges to positive infinity as $\beta\to0$, and (ii) the determinant of $\vec{\Sigma}_{\phi,\sigma^2}$, appearing in the RHS, remains bounded above.

In relation to (i), we derive a lower bound for the mutual information as follows:
\begin{align}
    \mathcal{I}(\vx,\vx^\prime)
    & =\KL(p_{\mathcal{D}}(\vx)p_{\theta,\phi}(\vx^\prime|\vx)||p_{\mathcal{D}}(\vx)p_{\theta,\phi}(\vx^\prime))\notag\\
    &=\mathbb{E}_{p_{\mathcal{D}}(\vx)p_{\theta,\phi}(\vx^\prime|\vx)}[\log p_{\theta,\phi}(\vx^\prime|\vx)-\log p_{\theta,\phi}(\vx^\prime)] \notag\\
    &=\mathcal{H}[p_{\theta,\phi}(\vx^\prime)]-\mathbb{E}_{p_{\mathcal{D}}(\vx)}\mathcal{H}[p_{\theta,\phi}(\vx^\prime|\vx)]\notag\\
    &\geq\mathcal{H}[p_{\theta,\phi}(\vx^\prime)] -\mathbb{E}_{p_{\mathcal{D}}(\vx)}H(\beta\vec{I}),
\end{align}
where $p_{\theta,\phi}(\vx^\prime|\vx) := \mathbb{E}_{q_\phi(\vz|\vx)}[p_{\theta}(\vx^\prime|\vz)]$ and $p_{\theta,\phi}(\vx^\prime) := \mathbb{E}_{p_{\mathcal{D}}(\vx)}[p_{\theta,\phi}(\vx^\prime|\vx)]$. According to Theorem~4 in \citep{dai2019diagnosing}, we have $\mathcal{H}[p_{\theta,\phi}(\vx^\prime)] \to \mathcal{H}[p_{\mathcal{D}}(\vx)]$ ($\beta\to0$), and since $H(\beta\vec{I}) \to -\infty$ as $\beta \to 0$, it follows that $\mathcal{I}(\vx,\vx^\prime) \to \infty$ as $\beta \to 0$.

In relation to (ii), let $V$ and $L$ represent constants corresponding to the maximum singular value of the covariance of $p_{\mathcal{D}}(\vx)$ and Lipschitz constant of $f_{\phi}(\vx)$. Since the additive noise for the reparameterization trick follows a normal distribution $\mathcal{N}(0,\vec{I}_d)$, the covariance matrix satisfies $\vec{\Sigma}_{\phi,\sigma^2}=\sigma^2\vec{I}+\vec{\Sigma}_{\phi}^e$. Therefore, we have
\begin{align}
    \det(\vec{\Sigma}_{\phi,\sigma^2})=\det(\sigma^2\vec{I}+\vec{\Sigma}_{\phi}^e)\leq(\sigma^2+VL^2)^d<\infty,
\end{align}
where the upper bound is derived using the fact that the maximum singular value of $\vec{\Sigma}_{\phi}^e$ is bounded above by $VL^2$.

Finally, applying (i) and (ii) to \Cref{eq:ineq_main} conclude this proof.
\end{proof}

\subsection{Proposition~\ref{prop:largebeta}}

\noindent\textit{Proposition}~\ref{prop:largebeta}~\textit{The mutual information between the input and the reconstructed samples produced by the VAE becomes zero as $\KL(q_\phi(\vec{z}|\vec{x};\beta=1)|| p(\vz))$ converges to zero. It holds for any decoder function $g_{\theta}(\cdot,\beta=1)$.}

This claim is almost direct consequence of Theorem~1 in \citet{takida2022preventing}. For the sake of completeness, we provide the proof in this manuscript.
\begin{proof}
Let $\bm{x}$ denote the input sample, with its corresponding latent variable represented as $\bm{z}$, and let $\bm{x}^\prime$ denote the reconstructed sample obtained via $\bm{z}$. These are defined as $\bm{z} \sim q_\phi(\bm{z}|\bm{x};\beta=1)$ and $\bm{x}^\prime \sim p_\theta(\bm{x}^\prime|\bm{z};\beta=1)$, respectively. Applying the data processing inequality to the Markov chain $\bm{x} \to \bm{z} \to \bm{x}^\prime$ yields:
\begin{align} \mathcal{I}(\bm{x};\bm{z}) \geq \mathcal{I}(\bm{x};\bm{x}^\prime), \label{eq:ineq_mi_x_z_1}
\end{align}
Additionally, the mutual information $\mathcal{I}(\bm{x};\bm{z})$ can be rewritten using its definition as follows:
\begin{align} \mathcal{I}(\bm{x};\bm{z}) &= \KL(p_{\mathcal{D}}(\bm{x})q_{\phi}(\bm{z}|\bm{x}) || p_{\mathcal{D}}(\bm{x})q_{\phi}(\bm{z}))\notag\\ &= \E_{p_{\mathcal{D}}(\bm{x})q_{\phi}(\bm{z}|\bm{x})}[\log q_{\phi}(\bm{z}|\bm{x}) - \log q_{\phi}(\bm{z})]\notag\\ &= \E_{p_{\mathcal{D}}(\bm{x})}\KL(q_{\phi}(\bm{z}|\bm{x}) || p(\bm{z})) - \KL(q_{\phi}(\bm{z}) || p(\bm{z})),
\end{align}
where $p(\bm{z})$ denotes the prior distribution over $\bm{z}$.
Since the Kullback-Leibler divergences $\KL(q_{\phi}(\bm{z}|\bm{x}) || p(\bm{z}))$ and $\KL(q_{\phi}(\bm{z}) || p(\bm{z}))$ are non-negative, it follows that: 
\begin{align} \mathcal{I}(\bm{x};\bm{z}) \leq \E_{p_{\mathcal{D}}(\bm{x})}\KL(q_{\phi}(\bm{z}|\bm{x}) || p(\bm{z})).
\label{eq:ineq_mi_x_z_2}
\end{align}
Combining inequalities~\eqref{eq:ineq_mi_x_z_1} and \eqref{eq:ineq_mi_x_z_2} completes the proof.
\end{proof}

%% file: app/experiment_deets.tex
\section{Details of experiments}
\label{app:exp_deets}
\subsection{Details of Implementation}
We will open-source the Python code used to implement the method described in this work, along with the supporting scripts used for running the experiments. Our implementation is primarily based on the code provided by \citet{Rombach2021HighResolutionIS}. For our method, we extend this existing codebase to implement our multi-$\beta$ VAE (\cref{subsec:multibeta-VAE}), the interpolating objective function based on a VAE's ELBO objective (\cref{eq:condbetaelbo}), and the non-linear diffusion model (\cref{subsec:non_linear_dm}). Three Python files containing these core components of our implementation are included with our submission. A detailed list of the hyperparameters used in the experiments is provided below.

\subsubsection{Disentanglement on Face Images}

\begin{table}[h]
\centering
\caption{Multi-$\beta$ VAE architecture configuration for TAD measurement on CelebA.}
\label{tab:tad_multibetaVAE}
\begin{tabular}{|l|c|}
\hline
\textbf{Parameters} & \textbf{CelebA} \\
\hline
Base channels & 32 \\
Channel multipliers & {[1, 2, 4]} \\
Attention resolutions & None \\
Number of res. blocks & 2 \\
Latent channels & 4 \\
Resolution & 64 \\
Input / Output channels & 3 / 3 \\
Dropout & 0.0 \\
Double $\vz$ & False \\
Embedding dim & 4 \\
Loss function & LPIPS + GAN + KL \\
\hline
\end{tabular}
\end{table}

\begin{table}[h]
\centering
\caption{Non-linear Diffusion model architecture for TAD measurement on the CelebA dataset.}
\label{tab:nonlin_diffusion_TAD}
\begin{tabular}{|l|c|}
\hline
\textbf{Parameters} & \textbf{CelebA} \\
\hline
Base channels & 64 \\
Channel multipliers & {[1, 2, 4, 4]} \\
Attention resolutions & {[1, 2, 4]} \\
Attention heads & 8 \\
Model channels & 64 \\
$\beta$ scheduler & Linear \\
$T=B$ & 1000 \\

\hline
\end{tabular}
\end{table}

The architecture of our multi-$\beta$ VAE and nonlinear diffusion model is designed to match that of \citet{yang2023disdiff}, with details presented in \cref{tab:tad_multibetaVAE} and \cref{tab:nonlin_diffusion_TAD}, respectively. For evaluation, we use the Total AUROC Difference metric and the code provided by \citet{yeats2022nashae}.

\subsubsection{Disentanglement on Toy Experiments}

To ascertain the representations learned by our method in \cref{subsec:exp1}, we utilize small multi-$\beta$ VAE architectures and employ the same convolutional backbone for implementing the baselines. Unlike previous works, we do not incorporate feed-forward layers in the architecture of the encoders and decoders. We assess the convergence of the model using the reconstruction loss. 

For the architecture, all input images are resized to $64 \times 64 \times 3$, and we do not apply any data augmentation when training the multi-$\beta$ VAE. For all methods presented in \cref{table:dis_results}, we use a latent dimensionality of $32$. The encoder and decoder are symmetric networks that are fully convolutional. The encoder consists of four layers with $[64, 64, 128, 256]$ channels, respectively, and we apply \texttt{GroupNorm} after every convolutional layer, followed by a \texttt{Sigmoid} non-linearity. 

For training these models, we select a batch size of $150$ and continue training until convergence is reached for each model. We used one NVIDIA L40S GPU for training on all these toy datasets. We trained our VAE with 500 values of $\beta$ equally spaced within the range $[0, 1]$. We perform a sweep over all $\beta$ values in our model to identify the optimal latent representation for each dataset. Specifically, for Cars3D, we achieve the highest score at $\beta = 285/500$; for Shapes3D, at $\beta = 210/500$; and for MPI3D, at $\beta = 280/500$. \cref{table:dis_results} demonstrates that our model achieves disentanglement performance comparable to or surpassing that of other disentanglement models.

\subsubsection{Unconditional Image Generation}
For architectures used in~\cref{subsec:exp2}, we follow~\citet{Rombach2021HighResolutionIS} and use architectures present in their code base. For all data sets, we used an image resolution of $256\times256$ and compress the input by a factor of 8 leading to a latent dimensionality of $32\times32\times4$. We fix the multi-$\beta$ VAE architecture across datasets with the encoder and decoder each containing $4$ layers, each comprising $2$ ResNet~\citep{resnethe2016} blocks with $[128,\ 256,\ 512, \ 768]$ channels, respectively.  

\begin{center}
\begin{minipage}{0.8\textwidth}
\centering
\begin{algorithm}[H]

\caption{Forward Pass for ResNet Block with Time Embedding}
\label{algo:resblock}
\SetAlgoLined
\KwIn{Input \( x \), Temporal embedding \( temb \)}
\KwOut{Output \( h \)}

\hspace{20pt}$h \leftarrow x$ \\
\hspace{20pt}$h \leftarrow \texttt{GroupNorm}(h)$ \\
\hspace{20pt}$h \leftarrow \texttt{Sigmoid}(h)$ \\
\hspace{20pt}$h \leftarrow \texttt{Conv}(h)$ \\

\hspace{20pt}$h \leftarrow h + \texttt{Proj}(\texttt{Sigmoid}(temb))$\\

\hspace{20pt}$h \leftarrow \texttt{GroupNorm}(h)$ \\
\hspace{20pt}$h \leftarrow \texttt{Sigmoid}(h)$ \\
\hspace{20pt}$h \leftarrow \texttt{Dropout}(h)$ \\
\hspace{20pt}$h \leftarrow \texttt{Conv}(h)$ \\

\hspace{20pt}\Return{$x + h$}

\end{algorithm}
\end{minipage}
\end{center}

We present a simplified version of the ResNet block we use in~\cref{algo:resblock}. We use attention at a resolution of $16\times16$ and the encoder and decoder are symmetric. The total number of trainable parameters on the multilevel $\beta$ -VAE is $148$ Million.

For implementing the non-linear diffusion model (\cref{subsec:non_linear_dm}), we increase the U-Net's output channels to $8$ and leave the rest of the architecture unmodified. As our diffusion model produces two outputs and is designed to model the noise at every time step along with a difference predictor (\cref{eq:nonlin_model_mean}) we use more than $300$ Million parameters in our testing and report results with a network having $378$ Million trainable parameters.\cref{tab:ldmhyperparams} lists the hyperparameters we used for training the unconditional diffusion models for all data sets. Next, we elaborate on the learned noise schedules.

\begin{table}[ht]
    \centering
    \caption{Hyperparameters for the unconditional LDMs producing the numbers shown in~\cref{tab:evaluation}. All models trained on 4 NVIDIA H100 GPUs.}
    \footnotesize
    \label{tab:ldmhyperparams}
    \renewcommand{\arraystretch}{1.3}
    \begin{tabular}{>{\centering\arraybackslash}m{3cm}c c c}
        \toprule
        & \textbf{CelebA-HQ 256 $\times$ 256} & \textbf{FFHQ 256 $\times$ 256} & \textbf{LSUN-Bedrooms 256 $\times$ 256} \\
        \midrule
        \textit{f} & 4 & 4  & 4 \\

        $z$-shape & $32 \times 32 \times 4$ & $32 \times 32 \times 4$ & $32 \times 32 \times 4$ \\

        Diffusion steps & 1000 & 1000 & 1000 \\

        $\mathcal{N}_{\text{params}}$ & 378M & 378M & 378M \\

        Channels & 256 & 256 & 256 \\

        Depth & 2 & 2 & 2 \\

        Channel Multiplier & 1,3,5 & 1,3,5 & 1,3,5 \\

        Attention resolutions & 32, 16, 8 & 32, 16, 8 & 32, 16, 8 \\

        Head Channels & 32 & 32 & 32 \\

        Batch Size & 98 & 98 & 98 \\

        Iterations* & 120k & 180k & 300k \\

        Learning Rate & 5.6e-5 & 4.4e-5 & 4.4e-5 \\
        \bottomrule
    \end{tabular}
\end{table}

\paragraph{\textbf{Noise Schedule:}} As elaborated in~\cref{subsec:info_loss} our method learns $\sigma_\beta$ as a result of training the multi-$\beta$ VAE which is the first stage of training irrespective of the dataset. The $\sigma_\beta$ obtained at the termination of this first stage of training is dependent on the choice of schedules of $\beta$ which control the coefficients of \emph{reconstruction term} and \emph{regularisation term} in~\cref{eq:condbetaelbo}. In the following plot, we present two examples of learned noise schedules using empirically determined $\beta$ schedules and contrast it with a linear noise schedule.
\begin{figure}[ht!]
    \centering
    \includegraphics[width=0.5\linewidth]{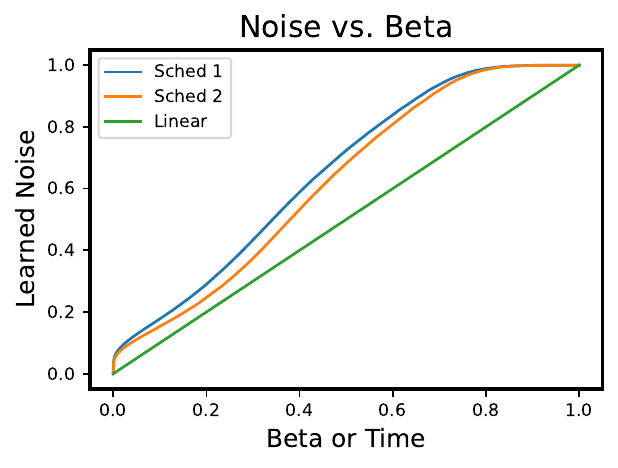}
    \caption{Examples of preferable noise schedules in our framework. These noise schedules have been used for learning non-linear diffusion models in~\cref{tab:evaluation}. While CelebAHQ~\citep{karras2018progressive} trains better with \texttt{Sched 1}, both FFHQ~\citep{karras19styleGAn} and LSUN-Bedrooms\citep{fisherLSUN} train better with \texttt{Sched 2}.}
    \label{fig:noisevbeta}
\end{figure}

The training of the non-linear diffusion model is sensitive to the setting  of this schedule, for instance, none of our models work well for a schedule that grows sub-linearly for $t<0.5$. The sudden increment in noise close to $\beta=0$ does not affect the model's performance negatively. Such a simga schedule is obtained by following a sinusoidal schedule for $\beta$. The exact form of our schedule is present in the shared code.

\subsection{Latent manipulation and interpolation}

\begin{center}
\begin{minipage}{0.9\textwidth}
\begin{algorithm}[H]
\label{algo:manipulation}
\DontPrintSemicolon
    \caption{Manipulate images with given direction}
    \SetKwInOut{Input}{Input}
    \SetKwInOut{Output}{Output}
    \
    \LinesNumbered
    
    \Input{Multi-$\beta$ VAE $(\vec{\theta},\vec{\phi})$, Non-linear diffusion model $(\vec{\epsilon}_\psi,\vec{\Delta}_\psi)$, an image $\vx$, latent direction $\vec{\delta}$, manipulation factor $\alpha$, time step $t=\beta$}
    \Output{Manipulated image $\vx_{\mathrm{man}}$}
    \BlankLine

    \hspace{20pt}Encode the image as $\vz_{\beta} = f_\phi(\vx,\beta)$\;
    
    \hspace{20pt}Add the perturbation to the encoded latent $\tilde{\vz}_\beta=\vz+\alpha\vec{\delta}$\;

    \hspace{20pt}Denoise $\tilde{\vz}_\beta$ back to $\beta=0$ using diffusion model $(\vec{\epsilon}_\psi,\vec{\Delta}_\psi)$\;
    
    \hspace{20pt}Decode $\tilde{\vz}_{0}$ as $\vx_{\mathrm{man}}=g_\theta(\tilde{\vz}_{0},\beta=0)$\;
    
    \hspace{20pt}\Return $\vx_{\mathrm{man}}$\;
\end{algorithm}

\begin{algorithm}[H]
\label{algo:interpolation}
\DontPrintSemicolon
    \caption{Interpolate a pair of images}
    \SetKwInOut{Input}{Input}
    \SetKwInOut{Output}{Output}
    \
    \LinesNumbered
    
    \Input{Multi-$\beta$ VAE $(\vec{\theta},\vec{\phi})$, Non-linear diffusion model $(\vec{\epsilon}_\psi,\vec{\Delta}_\psi)$, images $\vx_1$ and $\vx_2$, interpolation factor $\alpha$, target dimensions to interpolate $\Lambda$, time step $t=\beta$}
    \Output{Interpolated image $\vx_{\mathrm{int}}$}
    \BlankLine

    \hspace{20pt}Encode the images as $\vz_{\beta,1} = f_\phi(\vx_1,\beta)$ and $\vz_{\beta,2} = f_\phi(\vx_2,\beta)$\;
    
    \hspace{20pt}Initialize the latent variable as $\tilde{\vz}_{\beta}=\vz_{\beta,1}$\;

    \hspace{20pt}Interpolate latent variables as $\tilde{\vz}_{\beta}[j] = \texttt{Slerp}(\vz_{\beta,1}[j], \vz_{\beta,2}[j], \alpha)$ for $j\in\Lambda$\;
    
    \hspace{20pt}Denoise $\tilde{\vz}_\beta$ back to $\beta=0$ using diffusion model $(\vec{\epsilon}_\psi,\vec{\Delta}_\psi)$\;
    
    \hspace{20pt}Decode $\tilde{\vz}_{0}$ as $\vx_{\mathrm{int}}=g_\theta(\tilde{\vz}_{0},\beta=0)$\;
    
    \hspace{20pt}\Return $\vx_{\mathrm{int}}$\;
\end{algorithm}
\end{minipage}
\end{center}

In \cref{fig:architecture}, we obtain high-fidelity manipulated images by following Algorithm~\ref{algo:manipulation}. The candidates for the latent direction $\vec{\delta}$ are eigenvectors obtained via Principal Component Analysis (PCA) of a large set of latent embeddings. Algorithm~\ref{algo:interpolation} presents the pseudocode for interpolating between two images, with examples of the interpolated images shown in \cref{fig:interpolation_with_various_beta}. We begin by producing an interpolated encoding at a time step $t$ using an interpolation coefficient $\alpha$ to control the mixing between $\vz_1$ and $\vz_2$. Both algorithms require a trained multi-$\beta$ VAE and the corresponding nonlinear diffusion model to generate high-fidelity images.

%% file: app/extra_results.tex
\section{Additional Results}

This section presents additional results from applying our method to high-quality image datasets. In \cref{app:ours_vs_disco}, we demonstrate that our approach is complementary to learning-free methods by comparing it with DisCo~\citep{ren2022DisCo}. In \cref{app:beta_vs_mig_and_mi}, we investigate the relationship between disentanglement and the value of $\beta$. In \cref{app:more_reconstructions}, we show additional reconstructions generated by the multi-$\beta$ VAE. \cref{app:samples} includes samples generated by our model for qualitative evaluation, supporting the results presented in \cref{tab:evaluation} and confirming that our model can function as a standalone generative model. Furthermore, \cref{app:more_edits} explores the spectrum of learned latent representations through latent interpolation and attribute editing.

\color{c_yuhta}

\subsection{Orthogonality of our method and DisCo}
\label{app:ours_vs_disco}

In this subsection, we follow the same experimental setup as \cref{subsec:exp1_toydata}.
As shown in \cref{table:moreTab1}, our multi-$\beta$ VAE achieves better disentanglement performance compared to DisCo applied to standard VAE. Furthermore, DisCo consistently performs best when applied to multi-$\beta$ VAE, indicating that DisCo and our approach are complementary. Notably, while DisCo significantly improves standard VAE’s performance, its impact on multi-$\beta$ VAE is relatively marginal, suggesting that our model’s latent space is already well-disentangled.

\begin{table}[tb]
    \centering
    \sisetup{separate-uncertainty=true, table-format=+1.3(3), retain-zero-uncertainty=true, detect-weight, mode=text}
    \caption{
    Disentanglement results. The experimental setup follows Section 6.1, using identical VAE architectures for all methods.
    }
    \label{table:moreTab1}
    \renewrobustcmd{\bfseries}{\fontseries{b}\selectfont}
    \newrobustcmd{\B}{\bfseries}
    \resizebox{0.75\linewidth}{!}{%
    \begin{tabular}{lSSSSS}
         \toprule
         
         &{Metrics} & {VAE} & {DisCo on VAE} & {Multi-$\beta$ VAE} & {DisCo on Multi-$\beta$VAE} \\
         \midrule
         \multirow{2}{*}{\rotatebox{90}{Cars}} & {MIG ($\uparrow$)} & 0.037 \pm 0.022& 0.071 \pm 0.029 &   0.114 \pm 0.009 &  0.116 \pm 0.011\\
         
         &{DCI ($\uparrow$)}& 0.063 \pm 0.037 & 0.127 \pm 0.044 &  0.157 \pm 0.01 & 0.171 \pm 0.01\\
         \midrule
         \multirow{2}{*}{\rotatebox{90}{Shps}} & {MIG ($\uparrow$)}& 0.167 \pm 0.112 & 0.237 \pm 0.134 &  0.422 \pm 0.09 &  0.434 \pm 0.09 \\
         &{DCI ($\uparrow$)}& 0.363 \pm 0.027 & 0.526 \pm 0.045 &   0.621 \pm 0.09 &  0.678 \pm 0.081 \\
         \midrule
         \multirow{2}{*}{\rotatebox{90}{MPI}} & {MIG ($\uparrow$)}& 0.021 \pm 0.013 & 0.047 \pm 0.027 &  0.147 \pm 0.035 & 0.152 \pm 0.029 \\
         
         &{DCI ($\uparrow$)}& 0.096 \pm 0.024  & 0.188 \pm 0.042 &  0.253 \pm 0.043 &  0.284 \pm 0.060 \\
         \bottomrule
    \end{tabular}}
\end{table}

\subsection{The relationship between $\beta$ and disentanglement}
\label{app:beta_vs_mig_and_mi}

\begin{figure}[h]
    \centering
    \includegraphics[width=0.7\linewidth,trim= 10 15 0 10]{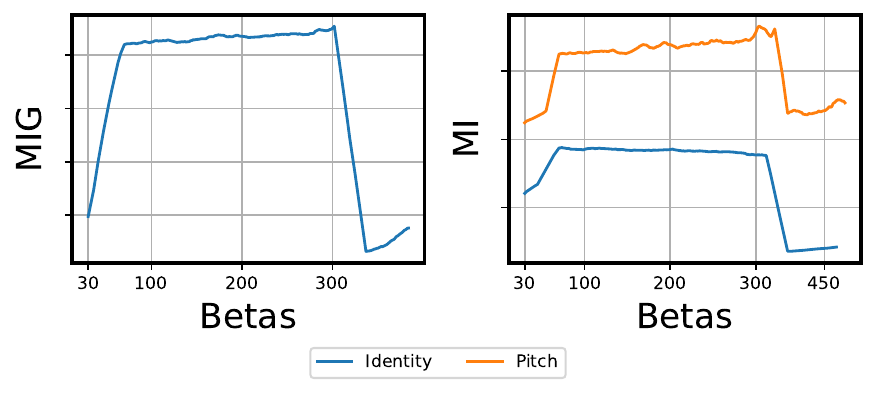}
    \caption{\textbf{Variation of MIG and MI:}
    We analyze the variation in MIG and MI scores of generative factors across different $\beta$ values on the Cars3D dataset~\citep{reedcars15} to quantitatively track disentanglement in the latent space. Notably, the highest MIG scores for each factor are achieved at different $\beta$ values.}
    \label{fig:plot_mig_beta}
\end{figure}

To further investigate the properties of latent representations learned at different $\beta$ values, we plot the evolution of the MIG score as a function of $\beta$ using the Cars3d dataset. These results are presented in \cref{fig:plot_mig_beta}. The plot reveals that MIG scores fluctuate in alignment with the loss function \eqref{eq:condbetaelbo} associated with each $\beta$ value. For models like $\beta$-TCVAE, $\beta$-VAE, and FactorVAE, determining an optimal coefficient for regularization can be challenging, often requiring numerous training runs as the ideal coefficient depends on both the dataset and model architecture. In our approach, this issue is mitigated by training a conditional $\beta$ model across a wide range of $\beta$ values (\cref{subsec:multibeta-VAE}).
Building on this analysis, we highlight an additional insight gained from using multiple $\beta$ values. By leveraging the distinct generative factors in the toy datasets, we plotted the highest mutual information for two factors across all $\beta$ values, \yuhta{showing that while mutual information (MI) increases for ‘Pitch’ as $\beta$ increases, it decreases for ‘Identity’ (see \cref{fig:plot_mig_beta})}. This observation supports the idea of training a model across multiple $\beta$ values, as it enables targeted manipulation of image factors by selecting specific $\beta$ values.

\color{black}

\subsection{Multi-$\beta$ VAE Reconstructions}
\begin{figure}
    \centering
    \includegraphics[width=\linewidth]{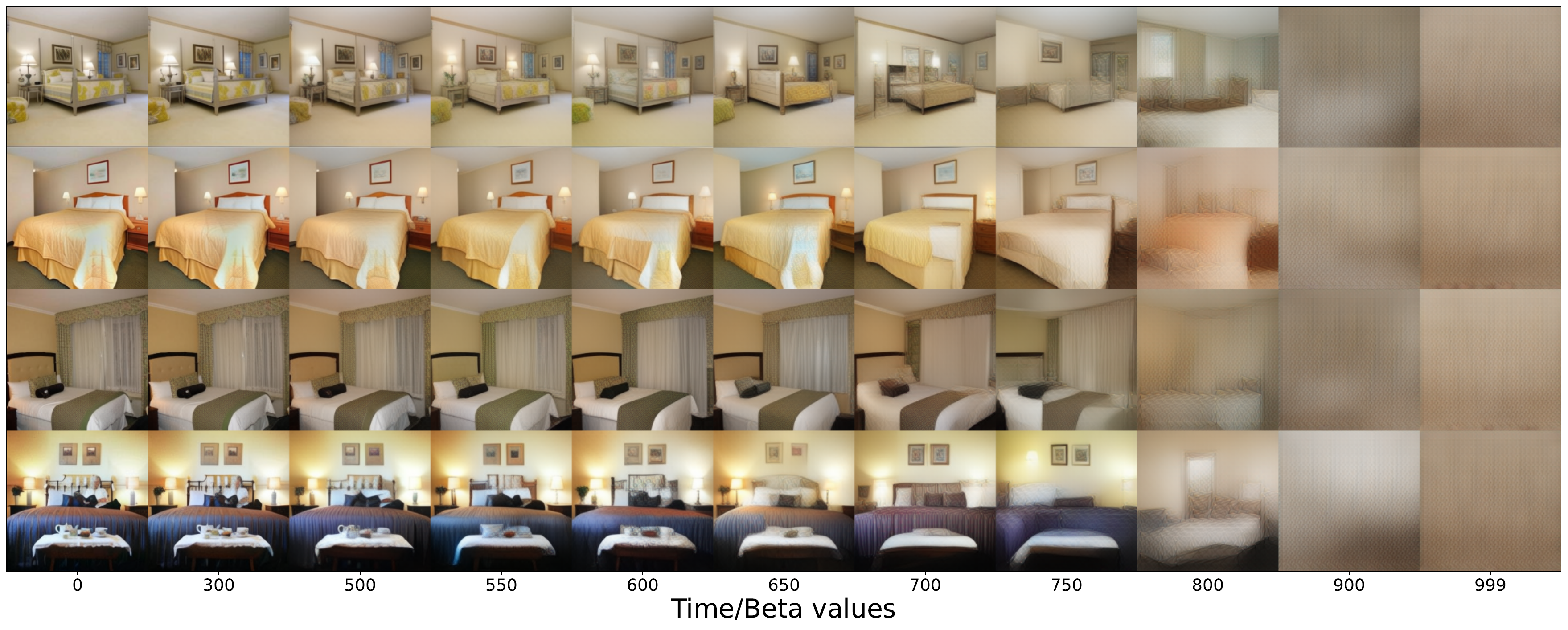}
    \newline
    \includegraphics[width=\linewidth]{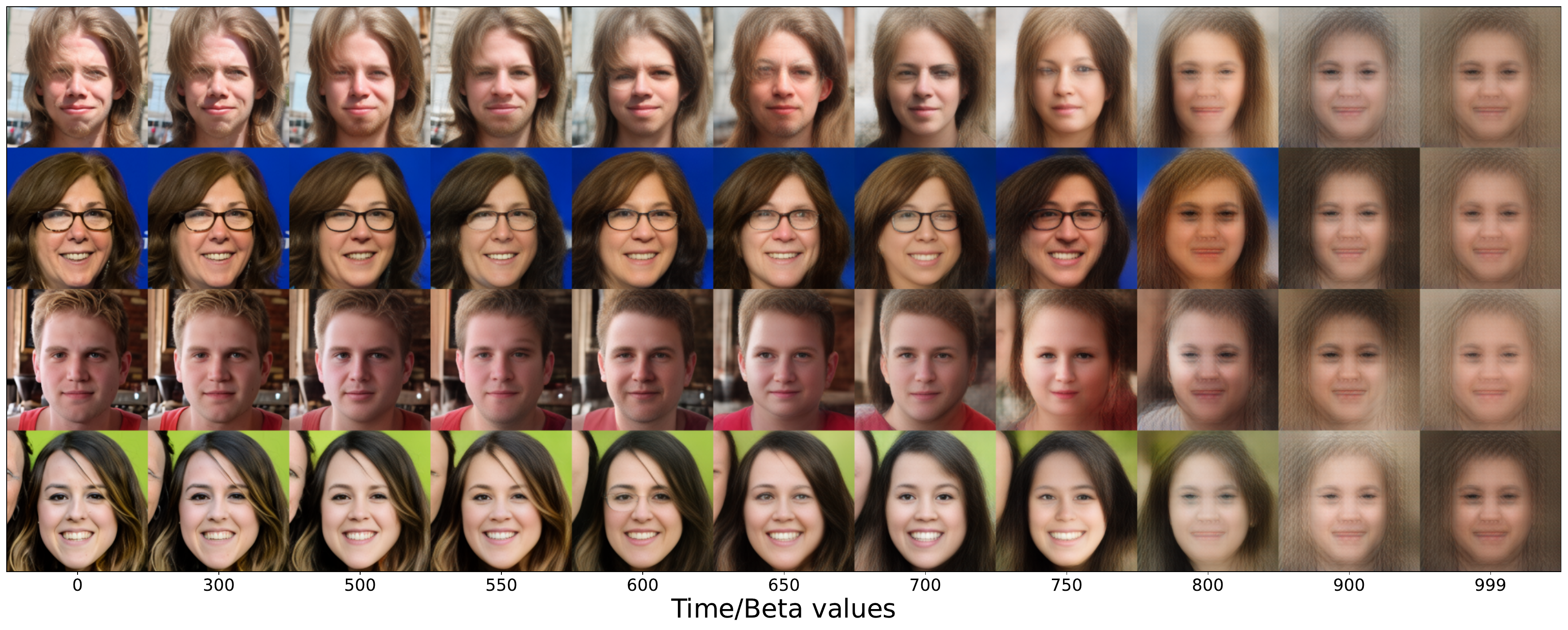}
    \newline
    \includegraphics[width=\linewidth]{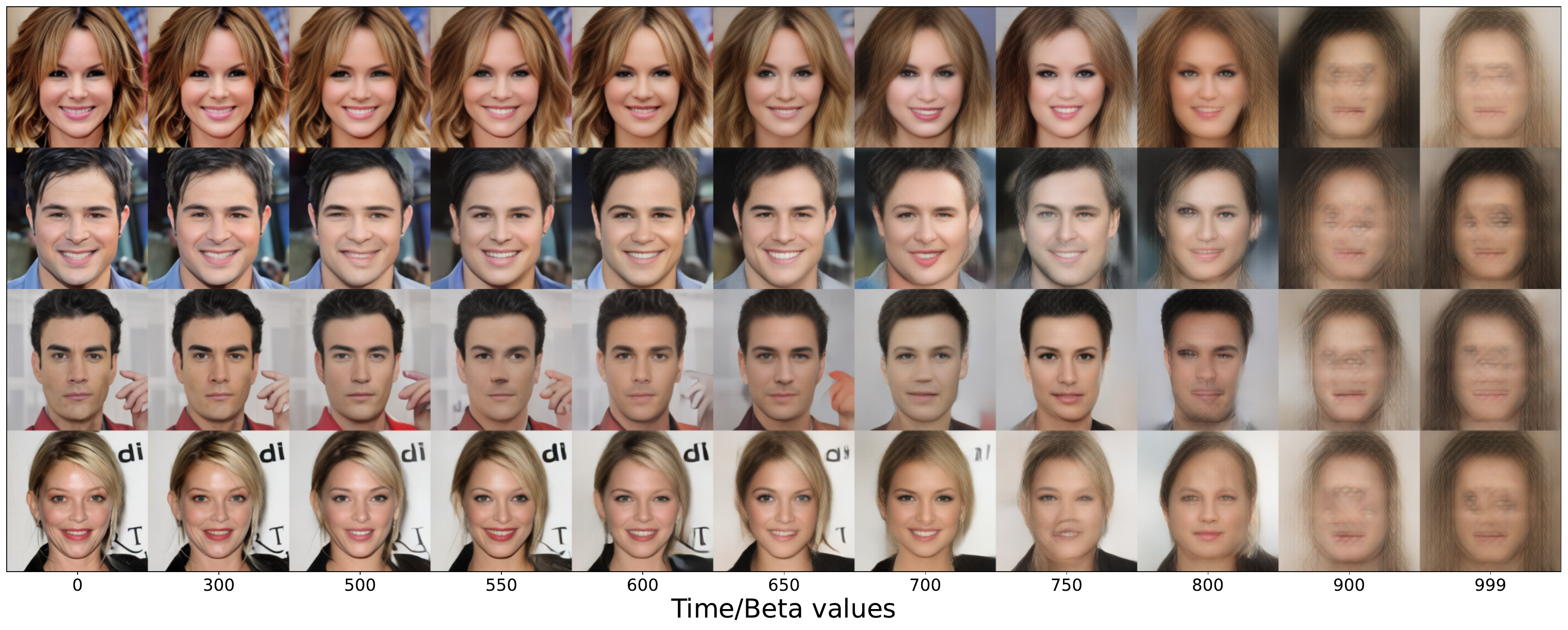}
    \caption{
    This figure presents examples of how our multi-$\beta$ VAE reconstructs images across different datasets as $\beta$ is varied. The top panel shows reconstructions on LSUN-Bedrooms~\citep{fisherLSUN}, the middle panel displays FFHQ~\citep{karras19styleGAn} reconstructions, and the bottom panel contains CelebA-HQ~\citep{karras2018progressive} reconstructions. Please refer to \cref{fig:smooth_latent} for reconstructed images through multi-$\beta$ VAE and non-linear diffusion model.
    }
    \label{fig:extra_recons}
\end{figure}

\cref{fig:extra_recons} shows reconstructions generated solely by our multi-$\beta$ VAE on the image datasets. To create these images, the same value of $\beta$ is passed to both the encoder and decoder networks. The figure displays a range of $\beta$ values, demonstrating that the latent representation loses information about the input image as $\beta$ increases. We would like to emphasize again that our non-linear diffusion model can generate additional information and decode the latent representation into high-fidelity images at any $\beta$ values, as shown in \cref{fig:smooth_latent}.

\label{app:more_reconstructions}
\subsection{Generated Samples from Noise}
\label{app:samples}
\begin{figure}
\centering

    \includegraphics[width=\linewidth]{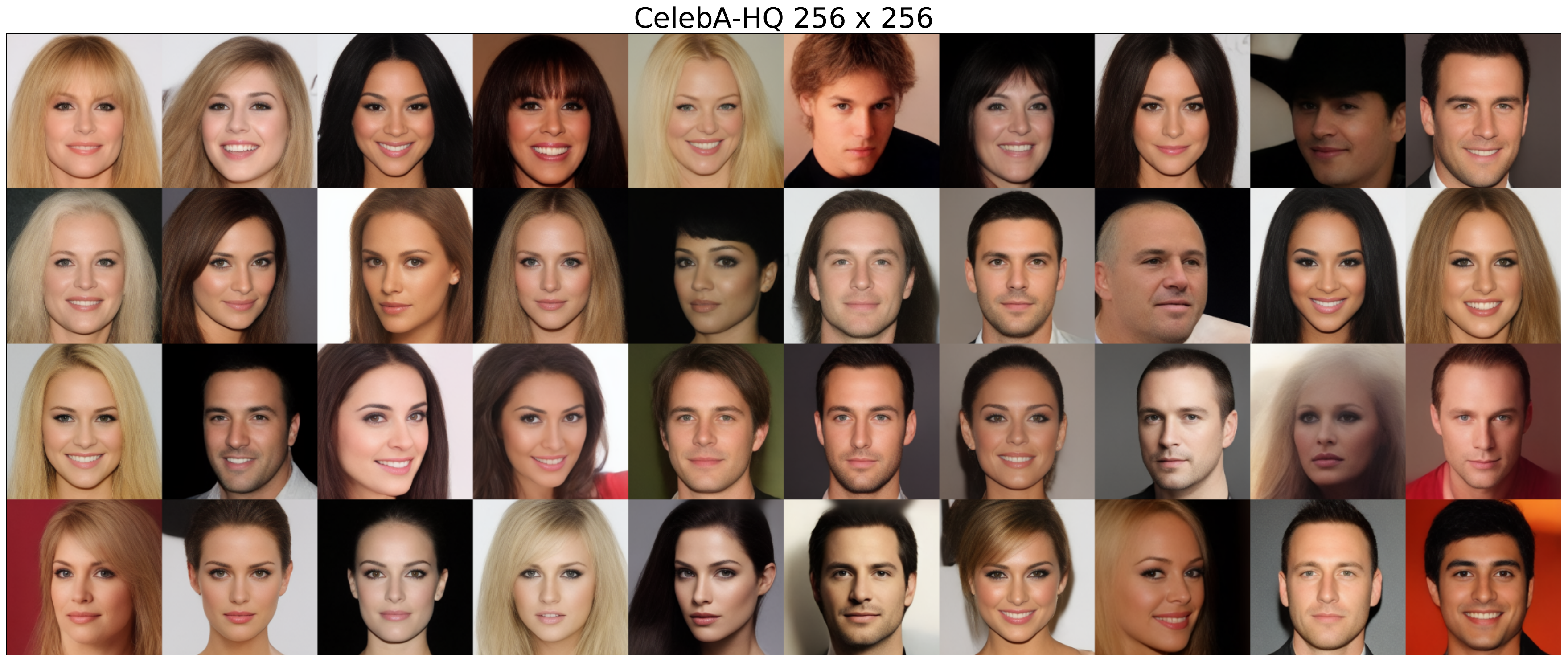}

    \includegraphics[width=\linewidth]{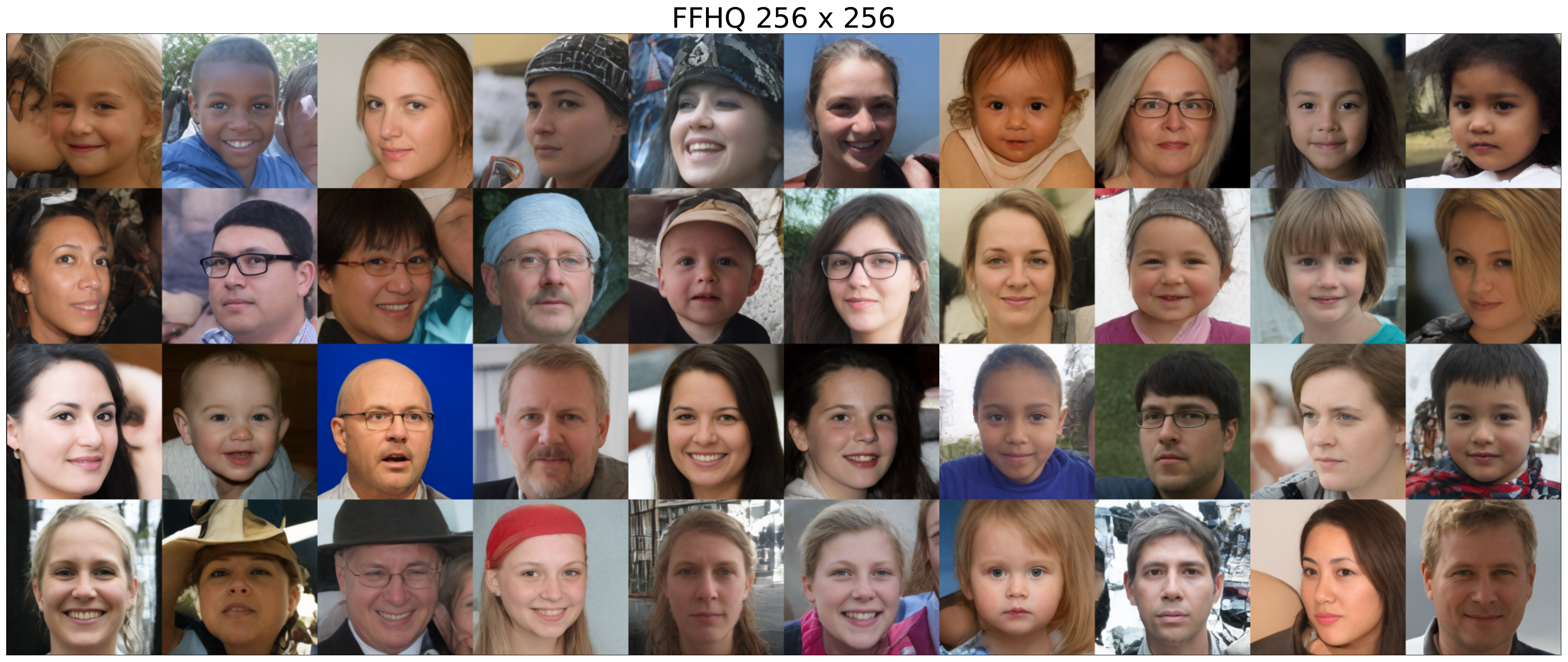} 

    \includegraphics[width=\linewidth]{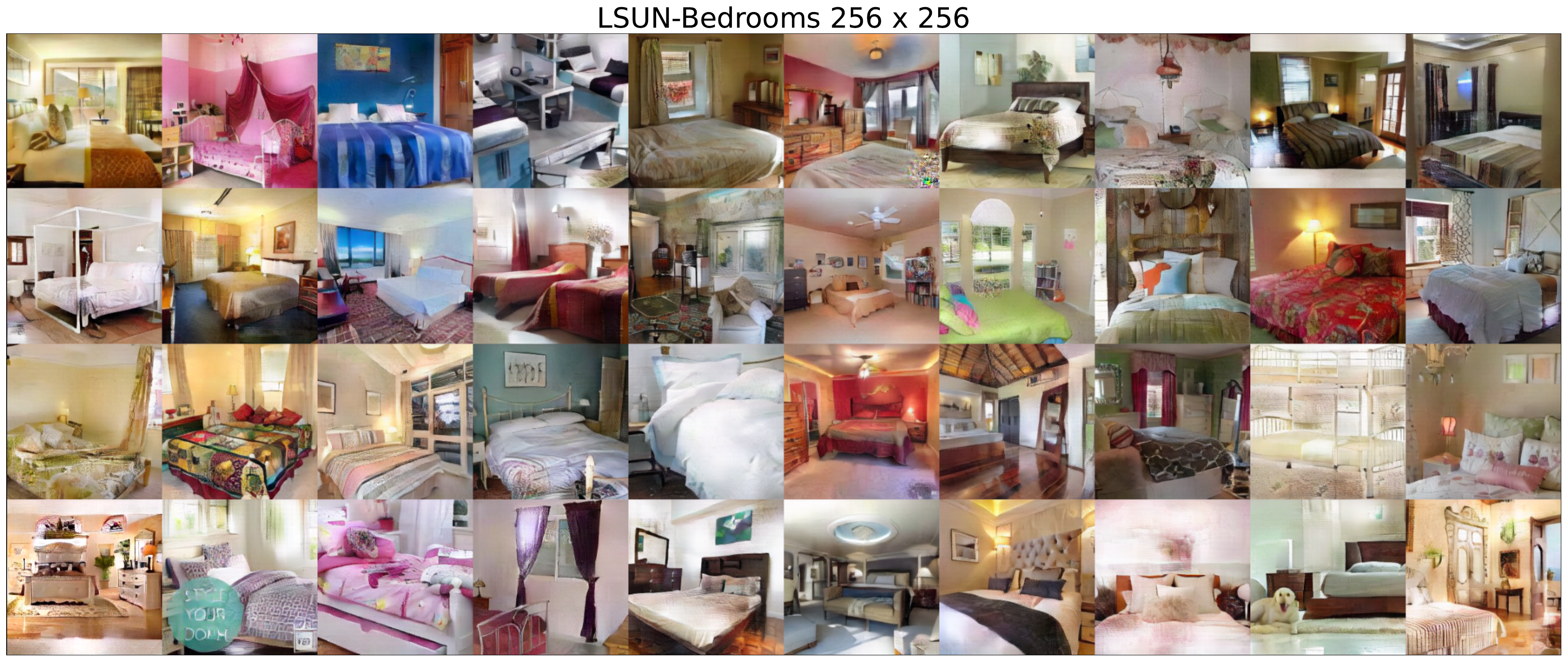}

\caption{
Samples generated by models trained on CelebA-HQ~\citep{karras2018progressive} (top), FFHQ~\citep{karras19styleGAn} (middle), and LSUN-Bedrooms~\citep{fisherLSUN} (bottom). These samples demonstrate that our model can function as a standalone generative model.
}%
\label{fig:ldm_samples}

\end{figure}

In \cref{fig:ldm_samples}, we show samples generated by our method, which combines a multi-$\beta$ VAE with a non-linear diffusion model, on the image datasets listed in \cref{tab:evaluation}. To generate samples, we begin by sampling a batch of noise from a standard normal distribution and then apply \cref{algo:sampling} to denoise it to $\beta=0$. In the final step, we decode the denoised latent sample into the pixel space using the trained decoder as $\vx = g_\theta(\vz_0,\beta=0)$. We train our non-linear diffusion models for $1000$ steps (refer to \cref{tab:ldmhyperparams}) and thus sample for $1000$ steps.

\subsection{Exploring the Spectrum of Learned Latents}
\label{subsec:exp4}

\begin{figure}[t]
    \centering
    \begin{subfigure}{\linewidth}
        \includegraphics[width=1.0\linewidth]{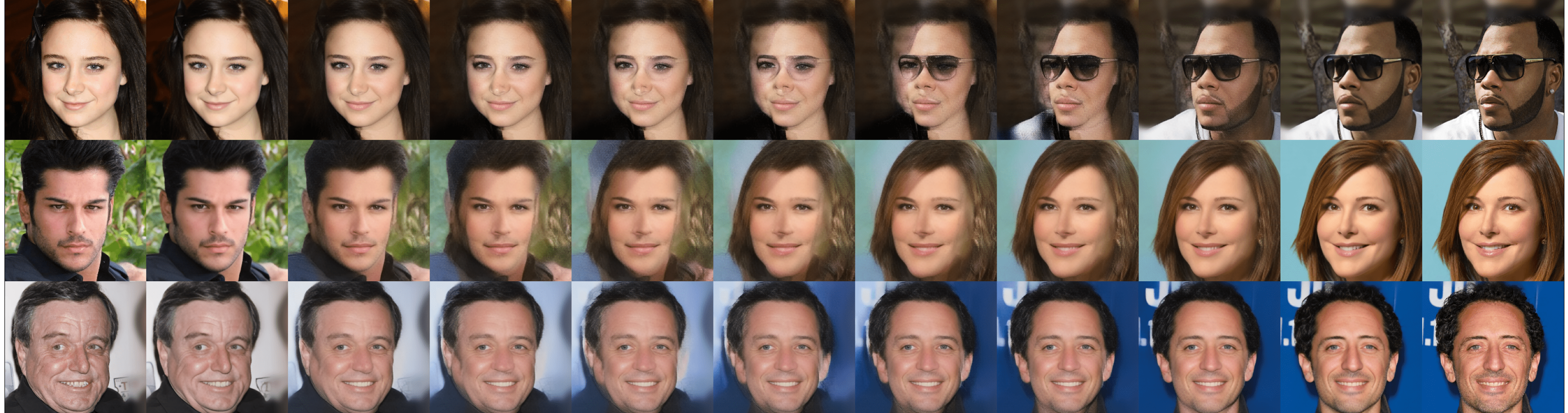}
        \caption{$\beta=0/1000$}
    \end{subfigure}
    \begin{subfigure}{\linewidth}
        \includegraphics[width=1.0\linewidth]{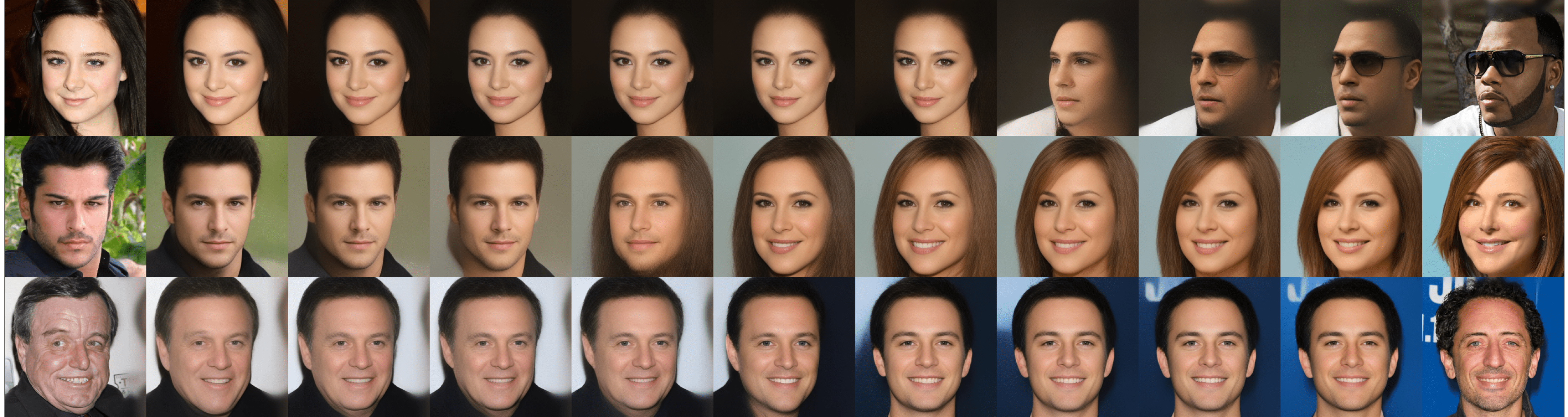}
        \caption{$\beta=300/1000$}
    \end{subfigure}
    \begin{subfigure}{\linewidth}
        \includegraphics[width=1.0\linewidth]{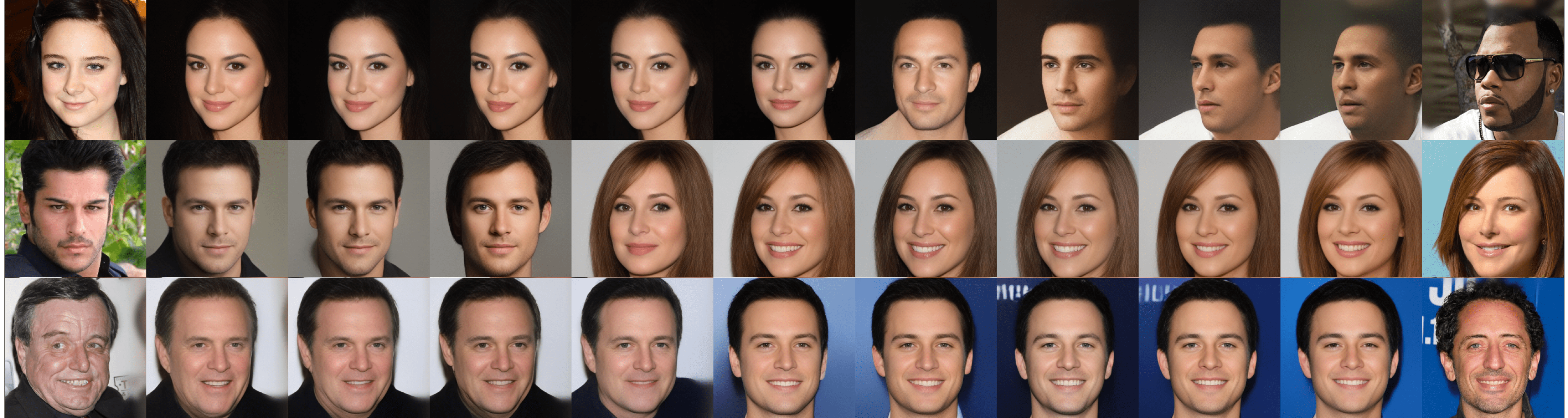}
        \caption{$\beta=600/1000$}
    \end{subfigure}
    \begin{subfigure}{\linewidth}
        \includegraphics[width=1.0\linewidth]{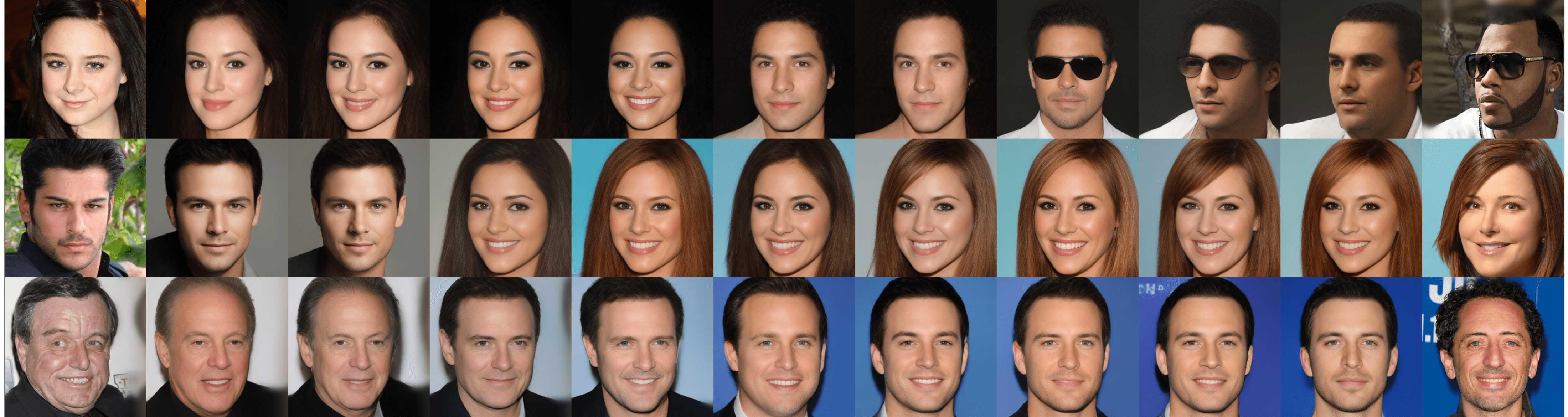}
        \caption{$\beta=900/1000$}
    \end{subfigure}
    \caption{\textbf{Interpolation in latent space:}
    We interpolate pairs of images (shown in the leftmost and rightmost columns) by applying slerp to the latent spaces.
    }
    \label{fig:interpolation_with_various_beta}
\end{figure}

\begin{figure}
    \centering
    \includegraphics[width=0.4\linewidth]{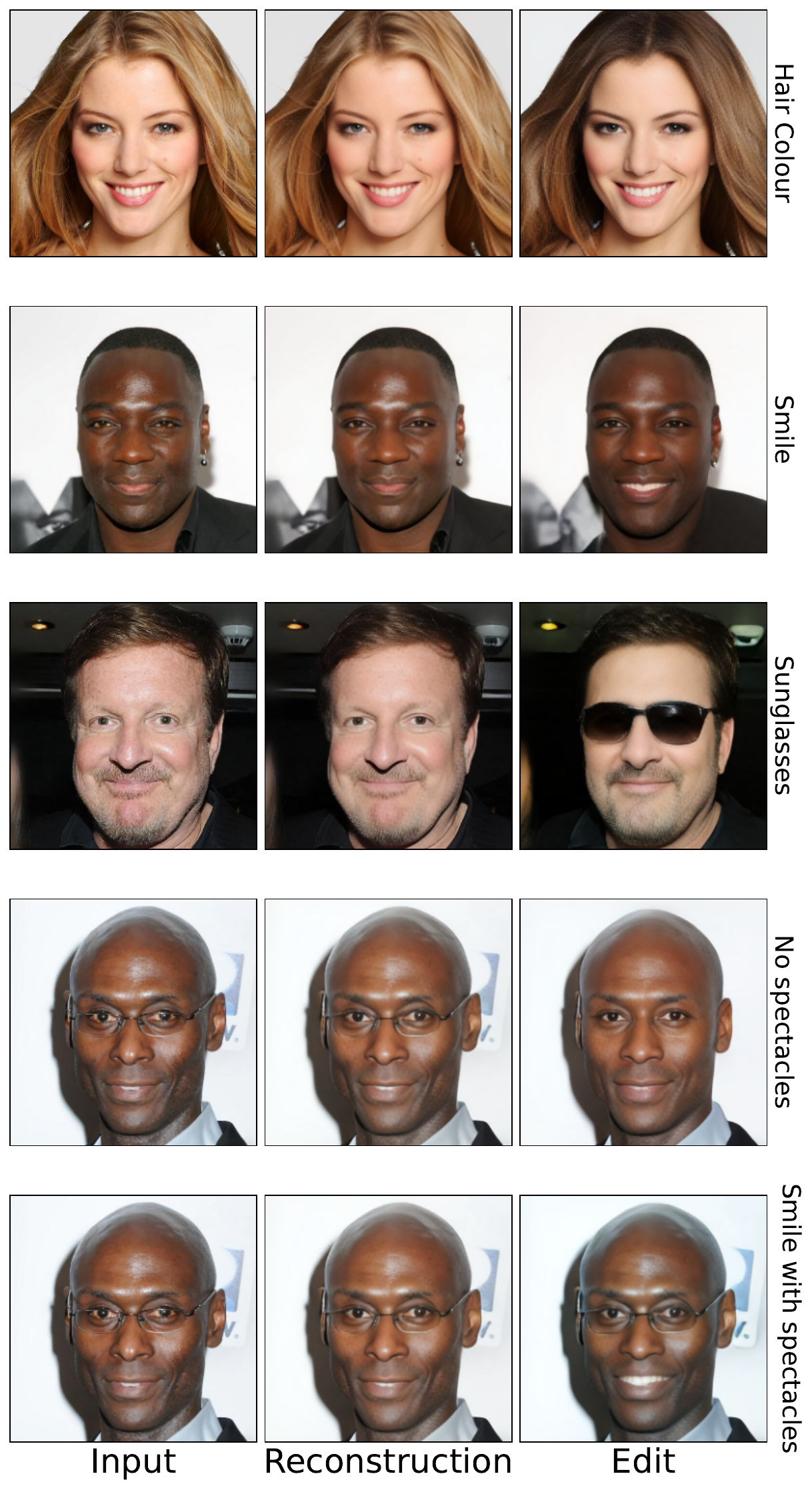}
    \caption{This figure contains more examples of editing of facial attributes of a person. In the first row we modify hair colour, in the second row we add a smile, in the third row we add sunglasses, in the fourth row we remove spectacles and in the final row we keep the spectacles while adding a smile. We source these input images from the CelebA-HQ~\citep{karras2018progressive} and FFHQ~\citep{karras19styleGAn} datasets.}
    \label{fig:editing_big_ex}
\end{figure}

We investigate the properties of latent spaces by interpolating pairs of images using spherical linear interpolation (slerp), as shown in \cref{fig:interpolation_with_various_beta}. We selected four values for $\beta$ to explore how the $\beta$ value affects the interpolated images. The latent space with the smallest value, $\beta=0$, is not regularized by the KL term, resulting in continuous but meaningless interpolation. As we gradually increase the value of $\beta$, the interpolated images transition more smoothly in the meaningful pixel space. Larger values of $\beta$ yield more natural-looking face images, although this comes at the cost of continuity in the image transitions. It is worth noting that even with large $\beta$ values, such as $\beta=900/1000$, the resulting images maintain high fidelity, which is a distinct property compared to traditional $\beta$-VAE. Additionally, we show examples of attribute changes in \cref{fig:editing_big_ex}. We follow Algorithm~\ref{algo:manipulation} to obtain the manipulated images.

\label{app:more_edits}

%% file: app/limitations.tex
\section{Limitations and Broader Impacts}
\label{app:limitations}

\paragraph{Limitations.}
Our model is trained in a completely unsupervised manner. While this approach allows us to learn disentangled representations without the need for labeled data, it necessitates the use of additional tools, such as PCA, to edit input images as desired, as demonstrated in the appendix. Although our primary focus is on developing generative models that can learn disentangled representations and perform downstream tasks without relying on extra conditioning, we recognize that incorporating text conditioning capabilities into our method could lead to more interpretable representations. This is a promising direction for future work.
Furthermore, our sampling method for the non-linear diffusion model is similar to DDPM rather than DDIM. Consequently, developing a DDIM-based sampling method for faster generation remains an area for future exploration.
Lastly, non-linear diffusion model is still a relatively unexplored area compared to linear diffusion models. While our model demonstrates promising results, we anticipate that future research focused on non-linear diffusion model could further enhance generation quality and expand the capabilities of generative models. 

\paragraph{Broader Impacts.}

As with any general image generation model, our approach carries the risk of producing harmful or inappropriate content, which is contingent upon the datasets used for training. We have utilized image datasets that have been widely accepted for academic purposes. However, it is essential to acknowledge that even well-curated datasets may contain biases or reflect societal stereotypes, which can inadvertently influence the outputs of our model.